\documentclass{article}
\usepackage{url}

\usepackage[ruled,vlined]{algorithm2e}
\usepackage{amsmath}
\usepackage{listings}
\usepackage{microtype}
\usepackage{graphicx}
\usepackage{subfigure}
\usepackage{booktabs} % for professional tables
\usepackage{xspace} % for \xspace command
\usepackage{hyphenat} % for \hyp command
\usepackage[dvipsnames]{xcolor}
\usepackage{minted} % for appendix code block
\usepackage{changepage}
\usepackage{appendix}
\usepackage{etoolbox}
\AtBeginEnvironment{minted}{\par\nolinenumbers}
\AtEndEnvironment{minted}{\par\linenumbers}
\usepackage{caption}
\setminted{
  frame=single,
  framesep=3mm,
  breaklines,
  breakanywhere,
  autogobble,
  fontsize=\footnotesize,
  baselinestretch=1.05,
  tabsize=2
}
\usepackage{dblfloatfix}

\usepackage{hyperref}

\usepackage[accepted]{mlsys2025}

\mlsystitlerunning{\fib: Building the Virtuous Cycle for AI-driven LLM Systems}

\newif\ifdraft
\drafttrue % for draft version
\AtBeginDocument{%
}

\newcommand{\fib}{FlashInfer-Bench\xspace}
\newcommand{\fitrace}{FlashInfer Trace\xspace}

\newcommand{\fibapplytt}{\texttt{flashinfer\_bench.}\texttt{apply()}}
\newcommand{\fibapplyenv}{\texttt{FIB\_ENABLE\_APPLY=1}\xspace}
\newcommand{\fibdata}{FlashInfer-Bench Dataset\xspace}
\newcommand{\fibapply}{\texttt{flashinfer\_bench.}\texttt{apply()}\xspace}
\newcommand{\fastp}{\ensuremath{\mathrm{fast}_{p}}\xspace}

\renewcommand{\algorithmiccomment}[1]{\{#1\}}

\makeatletter
\newcommand*\bigcdot{\mathpalette\bigcdot@{.7}}
\newcommand*\bigcdot@[2]{\mathbin{\vcenter{\hbox{\scalebox{#2}{$\m@th#1\bullet$}}}}}
\makeatother

\begin{document}

\twocolumn[
\mlsystitle{\fib: Building the Virtuous Cycle for AI-driven LLM Systems}

\mlsyssetsymbol{equal}{*}

\begin{mlsysauthorlist}
\mlsysauthor{Shanli Xing}{equal,uw}
\mlsysauthor{Yiyan Zhai}{equal,cmu}
\mlsysauthor{Alexander Jiang}{equal,cmu}
\mlsysauthor{Yixin Dong}{equal,cmu}
\mlsysauthor{Yong Wu}{nvidia}
\mlsysauthor{Zihao Ye}{nvidia}
\mlsysauthor{Charlie Ruan}{berkeley}
\mlsysauthor{Yingyi Huang}{cmu}
\mlsysauthor{Yineng Zhang}{independent}
\mlsysauthor{Liangsheng Yin}{independent}
\mlsysauthor{Aksara Bayyapu}{cmu}
\mlsysauthor{Luis Ceze}{uw,nvidia}
\mlsysauthor{Tianqi Chen}{cmu,nvidia}
\end{mlsysauthorlist}

\mlsysaffiliation{uw}{University of Washington}
\mlsysaffiliation{cmu}{Carnegie Mellon University}
\mlsysaffiliation{nvidia}{NVIDIA}
\mlsysaffiliation{berkeley}{University of California, Berkeley}
\mlsysaffiliation{independent}{Independent Researcher}

\mlsyscorrespondingauthor{Yixin Dong}{yixind@andrew.cmu.edu}
\mlsyscorrespondingauthor{Tianqi Chen}{tqchen@cmu.edu}

% Keywords for the paper
\mlsyskeywords{Machine Learning Systems, GPU Kernels, LLM Inference, Benchmark, Code Generation}

\vskip 0.3in

\begin{abstract}

Recent advances show that large language models (LLMs) can act as autonomous agents capable of generating GPU kernels, but integrating these AI-generated kernels into real-world inference systems remains challenging. \fib addresses this gap by establishing a standardized, closed-loop framework that connects kernel generation, benchmarking, and deployment. At its core, \fitrace provides a unified schema describing kernel definitions, workloads, implementations, and evaluations, enabling consistent communication between agents and systems. Built on real serving traces, \fib includes a curated dataset, a robust correctness- and performance-aware benchmarking framework, a public leaderboard to track LLM agents' GPU programming capabilities, and a dynamic substitution mechanism (\texttt{apply()}) that seamlessly injects the best-performing kernels into production LLM engines such as SGLang and vLLM. Using \fib, we further evaluate the performance and limitations of LLM agents, compare the trade-offs among different GPU programming languages, and provide insights for future agent design. \fib thus establishes a practical, reproducible pathway for continuously improving AI-generated kernels and deploying them into large-scale LLM inference.
\end{abstract}
]

\printAffiliationsAndNotice{\mlsysEqualContribution}

% Include all sections
\section{Introduction}

The rapid advancement of Large Language Models (LLMs) has catalyzed a new era of computing, but their widespread deployment is increasingly constrained by the performance and cost of their underlying inference systems \cite{NEURIPS2024_SGLang, kwon2023vllm, tensorrt-llm, mlc-llm}. At the heart of these systems are GPU kernels that execute the core operations like attention, matrix multiplication, and sampling. Optimizing GPU kernels used in LLM inference systems requires
deep, expert-level engineering effort.

\begin{figure*}[h]
\centering
\includegraphics[width=0.8\linewidth]{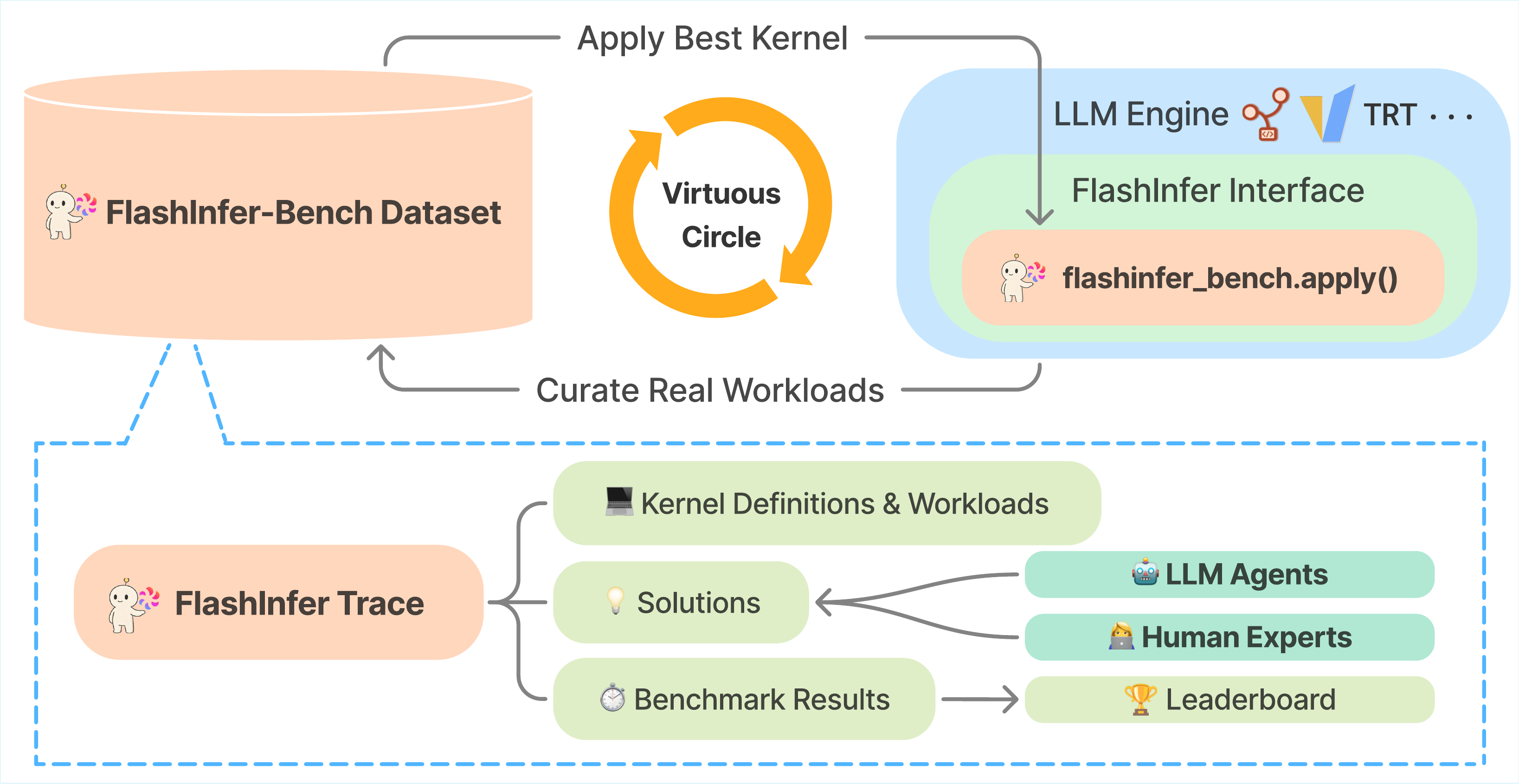}
\caption{\fib architecture. \fitrace provides a standard schema for specifying kernel contracts and semantics, and communicating implementations and evaluation results; \fibdata curates production LLM serving workloads; and \fibapply deploys the fastest validated implementation directly into LLM inference engines.}
\label{fig:fib-overview}
\end{figure*}

This paper asks a practical question: how can AI-generated kernels be effectively incorporated into production LLM systems? Recent works\cite{ouyang2025kernelbench,li2025tritonbench,baronio2025kevin,kernelllm2025} show early promise that LLMs can produce complex low-level GPU code. However, there are still three fundamental challenges to bridge AI generation to real-world deployments. First, kernels in LLM systems have many dependencies on different characteristics, such as ragged distribution, data precision, which impact their performance. It is important to effectively communicate this information to AI agents.
Second, real-world LLM inference traffic may differ from a typical uniform or random setup we pick in a single kernel. We need
an effective way to track the kernel performance on real-world LLM inference workloads. Finally, there still can be an
integration gap after AI-generating promising kernel candidates, as it can take extra engineering effort to bring them up to end-to-end LLM systems.

To address these challenges, we introduce \fib, a benchmark and standard operational flow for AI-driven LLM systems~(\autoref{fig:fib-overview}). To standardize overall workloads, we introduce \fib Trace, a self-contained standard JSON schema that describes the kernel task, workloads, the solution, and the final evaluation result.
Building on top of the schema, we curate the \fibdata from real-world LLM workloads. We also design a robust kernel benchmark framework on top of the \fitrace that features runtime isolation to prevent performance-related reward hacking and includes specialized support for evaluating low-bit and non-deterministic sampling kernels.
Finally, we built a dynamic kernel substitution mechanism to directly update the FlashInfer kernel library
to redirect operators to the optimal kernel provided by \fib trace in runtime.
This approach enables us to directly integrate common LLM-generated kernels into open source LLM engines such as SGLang and vLLM with no code change.

We build a live leaderboard to track the GPU programming capabilities of the frontier models
across real-world workloads and LLM workloads. We also did a comprehensive study of the current state of LLM agents
on real-world LLM inference systems.
Our evaluation and analysis show that:
(1) Most correctness errors come from compilation failures; (2) models struggle to exploit hardware-specific details such as architectural specifications or intrinsics; and (3) a language trade-off exists: high-level languages like Triton yield better performance on most tasks, while low-level CUDA provides more potential for specialized optimization.

The main contributions are as follows:

\begin{itemize}
\item We proposed \fitrace for standardizing the description of task, workload, and solution for AI-generated workloads.
\item We curated the \fibdata that serves a rich ground for evaluating AI-generated kernels on real-world workloads.
\item We proposed a pragmatic operational workflow to continuously generate and directly apply AI-generated kernels into a real-world production system.
\item We provided a comprehensive analysis of how LLM-generated kernels perform on LLM systems.
\end{itemize}

The rest of the paper is organized as follows: \autoref{sec:bg} reviews background on LLM inference, GPU kernels, and LLM for GPU kernel generation. \autoref{sec:design} presents the design of \fib, including the \fitrace schema, dataset curation, a robust performance-aware benchmarking framework, and dynamic kernel substitution for production engines. \autoref{sec:dataset} details the dataset and comprehensive evaluation of agent-generated kernels, with case studies on GEMM and GQA decode, as well as end-to-end substitution results. \autoref{sec:related_work} surveys related work. \autoref{sec:conclusion} concludes the paper.

\section{Background}
\label{sec:bg}

\subsection{LLM Inference Pipeline and GPU Kernels}

Modern LLM inference is powered by LLM serving engines, which handle batching, scheduling, and parallelism, and consist of GPU kernel invocations and CPU logic. GPU kernels dominates execution time, so optimizing them translates directly into reduced latency for the LLM engine. Despite model diversity, most models share a small set of GPU kernels, including:

\begin{enumerate}
	\item \textbf{GEMM}: Inputs and outputs can be bf16 or low-bit (e.g., fp8). It requires the use of tensor core instructions to achieve maximum speed. Low-bit variants require additional quantization/dequantization logic.
	\item \textbf{Attention and its variants}: E.g., paged, grouped, radix, and multi-latent attention. Requires tensor cores and needs special optimizations, such as FlashAttention, for its implementation.
	\item \textbf{Fused Mixture-of-Experts (MoE)}: A fused kernel that handles the MoE routing logic and multiple MLPs corresponding to multiple experts.
	\item \textbf{Sampling and post-processing}: E.g. top-p, top-k, temperature. These are non-deterministic operators whose results depend on the input distribution and random numbers.
\end{enumerate}

\subsection{Approaches to Kernel Optimization}

Kernel optimization is highly sensitive to hardware (SM count, memory hierarchy, tensor core generation), numerical format (FP16/BF16/FP8/INT8), and workload shape (sequence/batch lengths, cache layout, sparsity), making universal “one-size-fits-all” kernels elusive. System builders have relied on three families of techniques for kernel optimization:

\paragraph{Kernel libraries and templates.}
Highly optimized libraries provide strong baselines but often cannot exploit workload-specific structure (e.g., ragged sequences, fused epilogues) without custom kernels \citep{cutlass}.
\paragraph{Search-based auto-scheduling.}
Systems like template autotuners explore parameterized schedules within a fixed search space to find good tilings, thread/block mappings, and fusion strategies. They are powerful but bounded by the expressiveness of the template, and search costs can be prohibitive when the space must be revisited across hardware or shapes \citep{chen2018tvm,zheng2020ansor,tvm_metaschedule}.
\paragraph{Generative program synthesis.}
Recent LLMs can write low-level GPU code directly, sometimes discovering novel fusion and dataflow patterns beyond existing templates. This unlocks an enormous design space, but also introduces correctness and security risks without stringent validation; lightweight DSLs such as Triton make custom kernels accessible \citep{triton}.

\fib leverages the strengths of the last two: it enables generation to propose \emph{structurally new} kernels while surrounding it with a rigorous, production-grade evaluation harness to prevent regressions and reward hacking.

\subsection{LLM for GPU Code Generation}
Recent advances show that LLMs can synthesize non-trivial GPU kernels and fused operators when provided with an interface description and a feedback loop. Public evaluations such as KernelBench \cite{ouyang2025kernelbench} primarily assess \emph{generation capability}—can a model produce a compiling kernel that matches a reference on selected inputs and achieves reasonable speedups? In practice, moving from capability to \emph{production} demands additional ingredients: a precise task specification (API semantics, supported shapes/dtypes, memory layout constraints), defenses against reward hacking and non-determinism, coverage over realistic workload distributions, and an agile path to deploy or rollback candidates. Lightweight DSLs like Triton  \citep{triton} make authoring custom kernels accessible to both humans and agents, while \fib contributes the missing operational scaffolding—\emph{\fitrace} for standardized task exchange, a \emph{robust validator} to ensure safety and correctness, and \emph{day-0 dynamic substitution} to realize system-level gains without engine rewrites.

\section{\fib Design}
\label{sec:design}
\subsection{\fitrace}

\begin{figure}[h]
\centering
\includegraphics[width=0.9\linewidth]{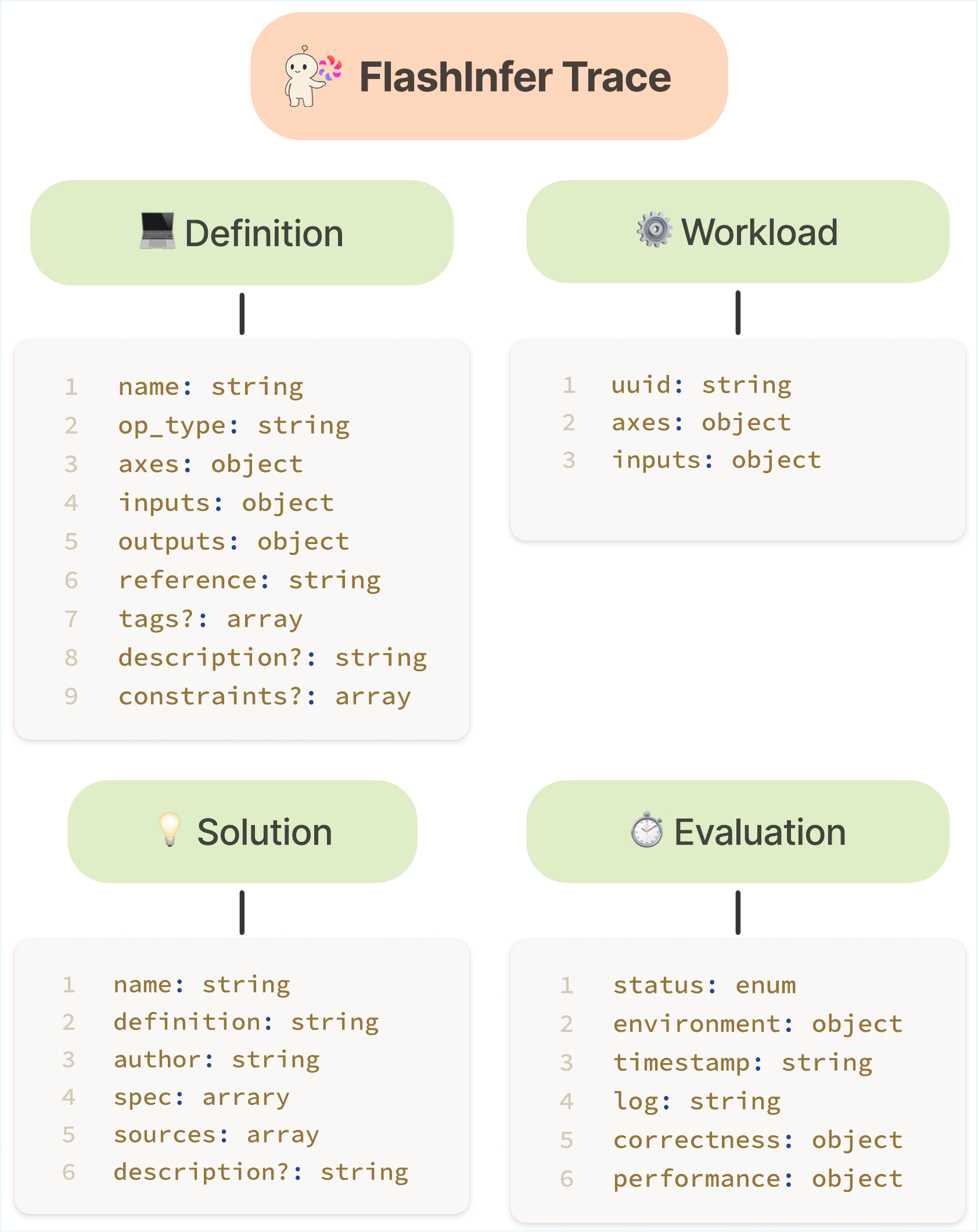}
\caption{\fitrace schema design. Definition describes the kernel task. Workload describes the real-world input to the kernel. Solution describes the AI-generated solution. Evaluation describes the evaluation result from benchmarking. Each component also includes auxiliary fields that are useful for grouping and filtering.}
\label{fig:fitrace}
\end{figure}

To close the loop from kernel generation to evaluation to deployment, we need a standardized language that is readable to both humans and agents. \fitrace serves as this common language. It articulates a kernel's semantic contract, implementation, and concrete evaluations. The abstraction is deliberately \emph{minimal} (e.g., we do not expose implementation-related system metadata in a kernel Definition) yet \emph{sufficient} (different operators introduce the key axes and constraints they need).

The \textbf{\fitrace schema} contains four components. An overview is shown in \autoref{fig:fitrace}. The four components below together form a self-contained \textbf{Trace} object, ensuring portability and reproducibility.

\begin{description}
    \item[\textit{Definition}.] A JSON specification of the operator's I/O tensors and dtypes, its dimension axes (could be a static value \textbf{const} or a workload-determined value \textbf{var}), and a plain \texttt{PyTorch}-based reference function \texttt{run} as the single source of mathematical semantics. Optional \texttt{constraints} encode relations among axes.
    \item[\textit{Workload}.] A concrete test input bound to a particular kernel Definition. All \emph{var} axes are assigned integer values, and each input is materialized via one of: recorded \texttt{safetensors} dump, \texttt{random} generation, or a literal \texttt{scalar} value.
    \item[\textit{Solution}.] A concrete implementation that satisfies a chosen Definition's interface and semantics. It provides source files and a callable entry point, alongside compatibility metadata (e.g., targeted GPU architectures and software versions). Extendable language/DSL support is included.
    \item[\textit{Evaluation}.] an immutable benchmarking record that precisely binds a specific \textbf{Definition} $\times$ \textbf{Solution} $\times$ \textbf{workload}, and reports the run state, a correctness and performance summary, and an execution-environment snapshot.
\end{description}

\fitrace schema natively supports dynamic and static kernel shapes, where each axis can be defined as either a \texttt{var} type (its value determined by the workload) or a \texttt{const} type (its value fixed at compile time). This enables AI to optimize kernels for specific shapes. It also supports ragged inputs, such as the page table used in Attention. To achieve this, both the full-page table tensor and the integer tensor storing index pointers can be provided as inputs, allowing the system to describe ragged tensor inputs precisely. We show concrete examples in \autoref{appendix:trace}.

\subsection{\fibdata}
\begin{figure}[h]
  \centering
  \includegraphics[width=0.9\linewidth]{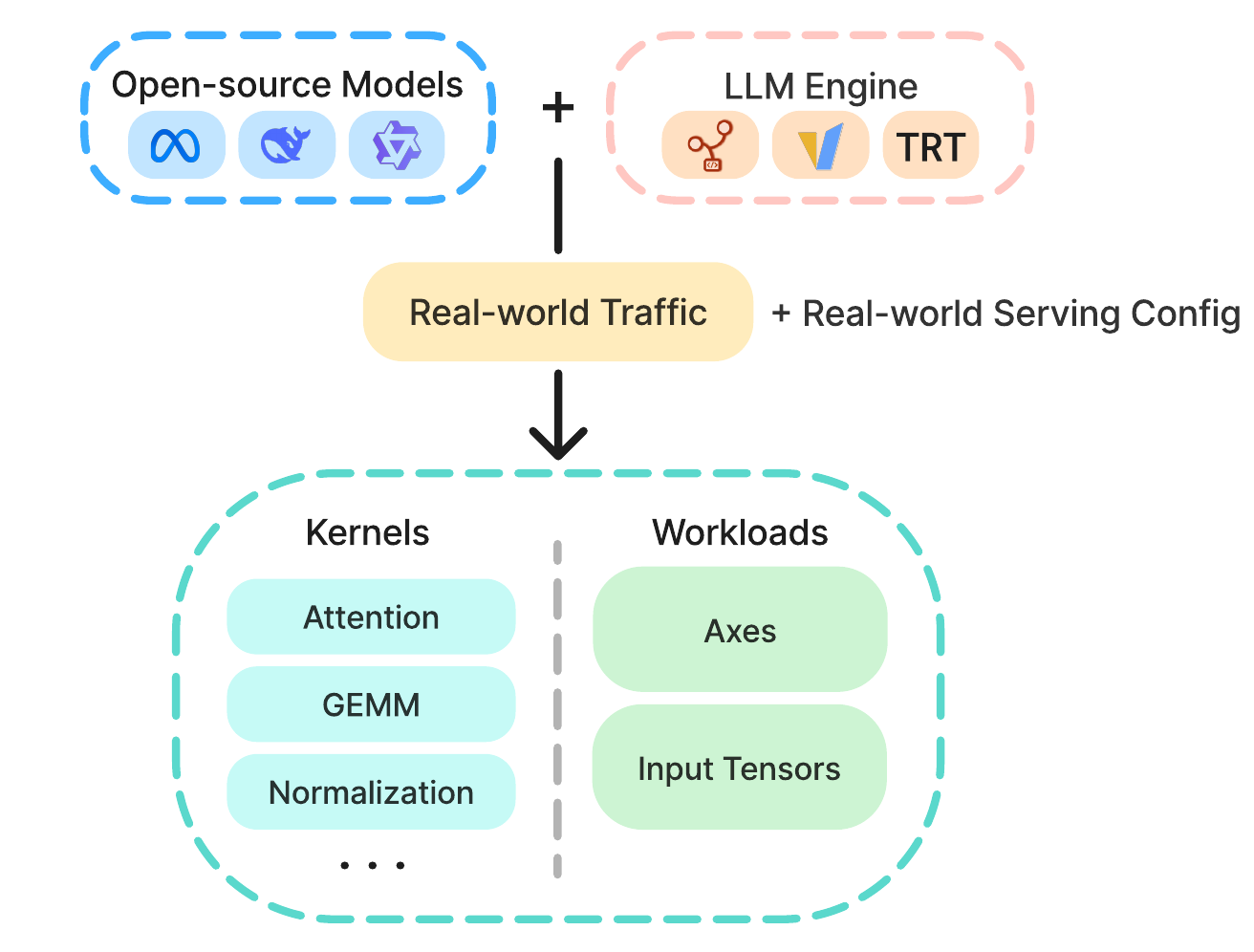}
  \caption{\fibdata collection workflow. We serve the major models against real-world traffic with default, common config, and curate the kernel definitions and workloads.}
  \label{fig:fibdata_curation}
\end{figure}

Building upon \fitrace, we curate the \fibdata, an \emph{evolving standard} that pairs common serving-kernel Definitions with representative Workloads, continually tracking and collecting targets that both humans and agents can optimize against.

Our objective is real-world relevance with a focus on LLM serving. We cover DeepSeek-V3, Llama-3.1-8B, Qwen3-30B-A3B across operator families, including GEMM, Attention, Normalization, Sampling, and MoE. Workloads are collected by running these models in SGLang with default, commonly used configurations (e.g., native FP8 quantization and tensor-parallel size of 8 for DeepSeek-V3) and serving them against ShareGPT prompts.

When collecting kernel Definitions, we classify two kernel invocations under the same Definition if and only if they (i) share the same I/O spec and \texttt{run} reference semantics; (ii) expose the same set of axes with identical const/var roles; and (iii) agree on all const-axis values.

We deliberately prefer specific Definitions over permissive ones, ideally down to a specific model layer kernel call, to enable best-effort kernel optimization and unambiguous dispatch. We disallow optional inputs or behavioral flags, and we encode default behavior inside the \texttt{run} reference so it becomes part of the contract. When behavior must diverge, we introduce a new Definition rather than adding runtime switches.

We curate and deduplicate the collected workloads with a \textbf{performance-aware, diversity-preserving} reduction. When input values materially affect kernel performance (e.g., sampling probability distributions) or correctness (e.g., extreme edge cases), we dump full tensors; otherwise, we use seeded random runtime tensors to save storage.

We then deduplicate along performance-sensitive axes (e.g., batch size) and tensor statistics (e.g., average sequence length for attention), trimming the set while preserving diversity and representativeness. In the end, each Definition keeps \(\sim 50\) workloads for evaluation.

\subsection{Robust Kernel Benchmark}
\label{sec:benchmark}

Targeting robustness and efficiency of the overall kernel evaluation process, we built a benchmarking subsystem that provides rigorous correctness validation and robust, reproducible timing, and runs natively in multi-device environments.

\paragraph{Deterministic Kernels.}
For operations expected to produce deterministic outputs (e.g., GEMM, normalization, attention, etc.), we directly compare the kernel’s output to the reference output elementwise. A kernel’s output $y_{\text{sol}}$ is considered correct if every element satisfies
\begin{equation}
|y_{\text{sol}} - y_{\text{ref}}| \leq \epsilon_{\text{abs}} + \epsilon_{\text{rel}} \cdot |y_{\text{ref}}|
\end{equation}
where $y_{\text{ref}}$ is the reference output. We also reject any output containing non-finite values (NaN or Inf). For each kernel, we record the maximum observed error across all tested elements and trials. A deterministic kernel passes the correctness check only if every output element falls under the permitted error bounds,.

\paragraph{Low-Precision Kernels.}
Kernels that use lower-precision arithmetic (e.g., FP8) introduce systematically larger errors compared to full-precision baselines. Instead of a single loosen global tolerance, we use a matched-ratio rule: a kernel is correct if at least $\rho$ of outputs meet the standard error criteria. For example, with $\rho = 0.95$, we require 95\% of the output elements to pass the tight error bounds, permitting a small percentage of outlier elements.

\paragraph{Stochastic Kernels.}

For stochastic operators like sampling, element-wise comparison is invalid since outputs vary per run. Instead, we verify that sampled outputs follow the correct probability distribution.
We derive the ground-truth distribution $\mathbf{q}$ from input probabilities $\mathbf{p}$ using an optional mask $\mathcal{M}$ (e.g., top-k or nucleus/top-p), normalize it, and repeatedly execute the kernel to obtain the empirical distribution $\hat{\mathbf{f}}$.
Finally, we compute the total variation distance (TVD) between the empirical and expected distributions and require $\text{TVD}$ to be smaller than a chosen threshold $\tau_{\text{TVD}}$. We use TVD because it directly upper-bounds the worst-case probability error over any event. In addition to the TVD check, we verify that every sample is accepted by the thresholding mask; any mask violation (e.g., sampling an index that should be excluded by the given top-k) results in immediate failure.

\paragraph{Performance Measurement.}
We maintain a per-GPU, multiprocess-visible device lock. To prevent interference among processes/tasks on the same device, the timing routine runs only after acquiring the device lock. Each kernel performs $w$ untimed warm-up runs followed by $m$ timed runs. We use CUDA event–based device-side timing and report the mean latency over the $m$ measured runs.

\paragraph{Isolation.}
To minimize cross-solution interference, and the risk of an LLM “hacking” the benchmark (e.g., by reading residual memory to infer reference outputs), we provide a fully isolated benchmarking mode: each solution runs in its own subprocess and is terminated on completion or timeout, with the CUDA context torn down to avoid cross-run state carryover. We also provide a persistent mode with one long-lived worker per GPU and a small spare pool of pre-warmed workers, which dramatically reduces subprocess and CUDA context initialization overhead and enables rapid recovery if a solution fails and corrupts a context. Together, these two modes balance the efficiency required for large sweeps with the robustness and safety guarantees of full isolation.

\paragraph{System Support.} As the dataset scales, efficient benchmarking becomes critical for our workflow to timely and sustainably operate. With that efficiency awareness, we build benchmarking a scalable, fault-tolerant multi-device service. For each ready \texttt{Solution×Workload} job, we have a scheduler that builds a cost matrix which accounts for baseline residency and warm compile caches against available device workers. It then assigns micro-batches with the Hungarian algorithm, then updates the cost model online via an exponentially weighted moving average before solving the next batch. The scheduler performs worker health checks and handles failure recovery. Execution defaults to persistent mode; solutions that repeatedly fail are deferred to isolated mode runs. Inputs and reference outputs are materialized on the target device and reused when possible. Compiled solutions are cached in memory and, when necessary, persisted to disk (e.g., CUDA binaries) to prevent redundant builds.

\subsection{Public Leaderboard and Continuous Evaluation}

\begin{figure}[h]
\centering
\includegraphics[width=\linewidth]{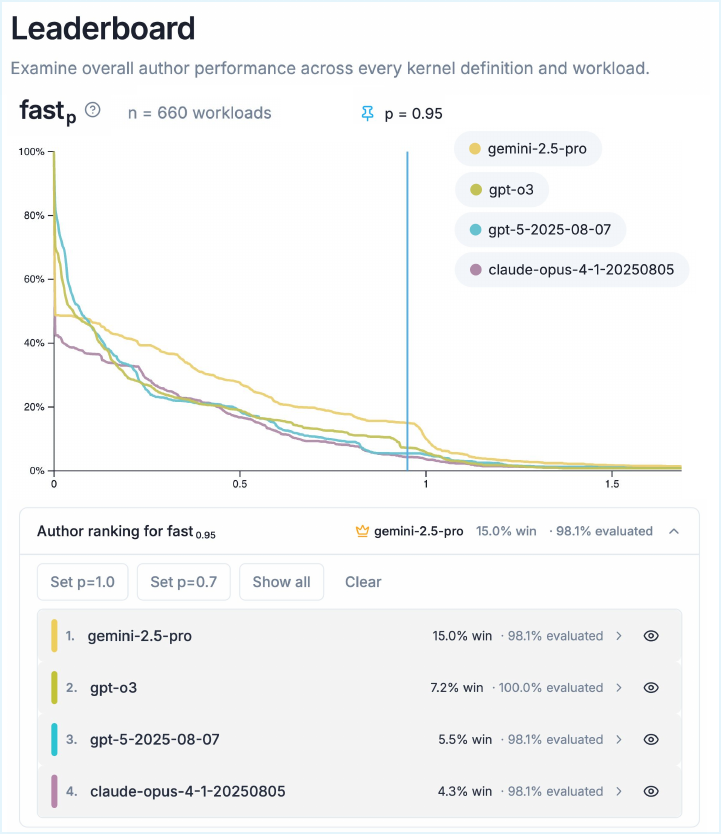}
\caption{FlashInfer-Bench Leaderboard. The top performing models at $\text{fast}_{0.95}$ are gemini-2.5-pro, gpt-o3, and gpt-5-2025-08-07. The top performing models in terms of correctness are gpt-5-2025-08-07 (83.9\% pass), gpt-o3 (71.3\% pass), and gemini-2.5-pro (48.8\% pass).}
\label{fig:web}
\end{figure}

We host a public leaderboard built on the benchmarking stack of \autoref{sec:benchmark} (see \autoref{fig:web}). It accepts submissions in the \fitrace format, evaluates them on real workloads, and reports kernel- and device-stratified metrics including correctness, performance curves versus speedup thresholds, per-workload latency, and end-to-end latency deltas. To ensure citability and reproducibility, we periodically release frozen snapshots with versioned datasets, while the rolling leaderboard reflects the latest evaluations. The service enforces anti-reward-hacking defenses (runtime isolation, hidden workloads, and dedicated validators for deterministic, low-precision, and sampling kernels). We analyze the current snapshot in \autoref{sec:dataset}.

\subsection{Dynamic Substitution of Kernels in Production}
Prior workflows require manual code changes inside the serving engine to deploy optimized kernels, creating a bottleneck for agent-driven automation and blocking an evaluation–deployment loop. To surface validated, optimized kernels and close this loop, we introduce \fibapply. It provides zero intrusion, low-overhead routing that dynamically maps serving requests to the best-performing implementation in the dataset.

\begin{figure}[h]
\centering
\includegraphics[width=0.48\textwidth]{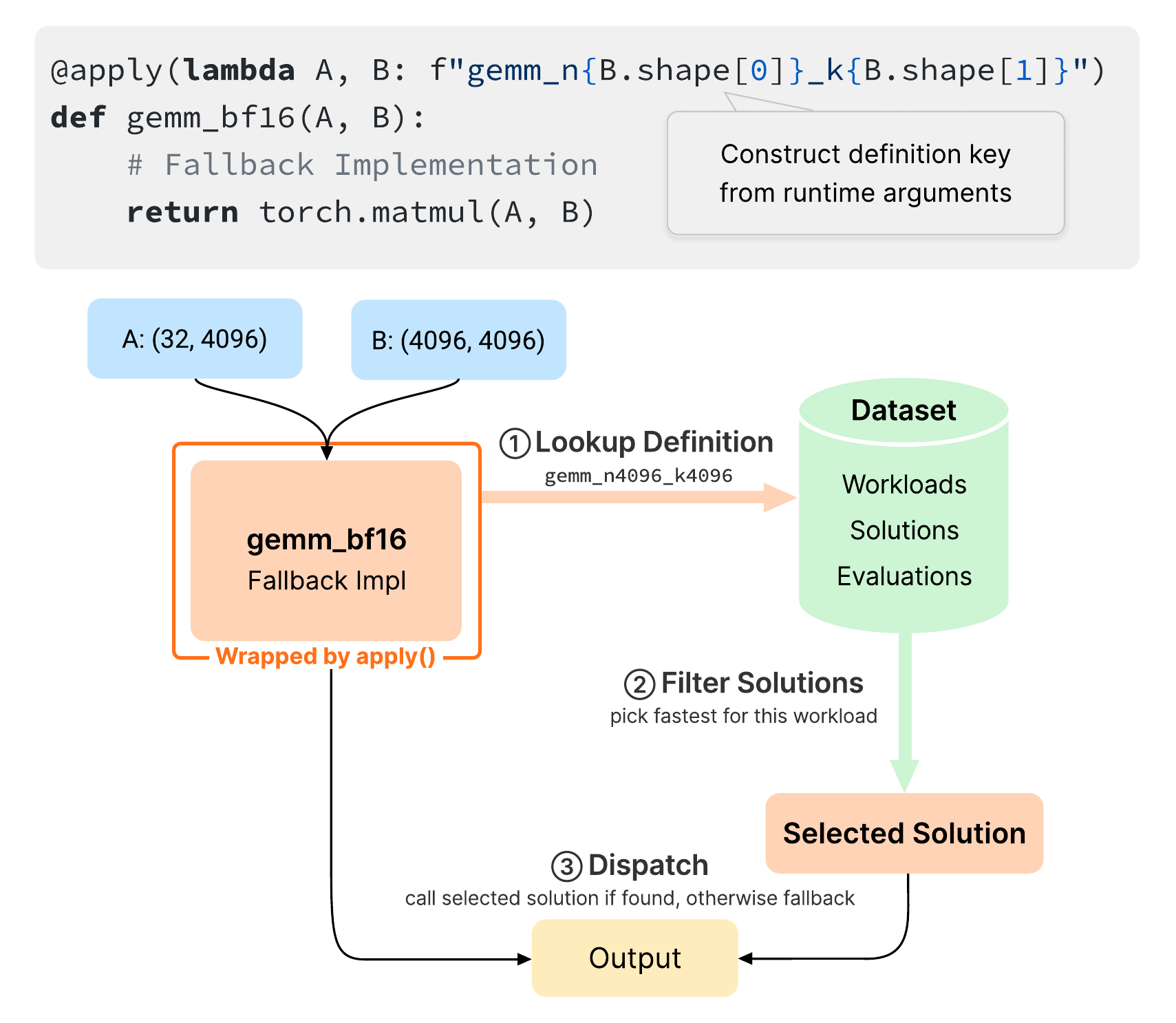}
\caption{Workflow of \fibapply. It is a dynamic dispatcher that retrieves the best Solution with the kernel input at runtime and returns the execution result.}
\label{fig:apply}
\vspace{-20pt}
\end{figure}

\paragraph{Interface and Usage.}
~\texttt{flashinfer\_bench.apply} \texttt{()} provides two APIs. The \textbf{decorator} API wraps an operator; the wrapped function serves as the fallback. It accepts either a fixed Definition name or a resolver that maps bound runtime arguments to a Definition name. The \textbf{imperative} API allows customized kernel invocation from arbitrary locations; it resolves the best-performing solution for the given inputs and returns the result immediately.  We provide first-class integration with FlashInfer, so common ops can be routed by enabling the environment variable \fibapplyenv without code changes.  When \fibapply is globally disabled, calls transparently pass through to the original implementation.

\paragraph{Offline Cache Prebuilding.}

To minimize \texttt{apply()}'s impact on serving performance, we introduce an ahead-of-time (AOT) built index for dynamic dispatch, reducing the online dispatch computation to a few \textbf{$O(1)$} index lookups.

Before the serving engine starts, the \texttt{apply()} runtime will initialize an index from the local dataset. It first filters traces by a configurable error threshold, then extracts features (e.g. shapes) from workloads in the traces to form a key, and for each key, picks the fastest solution as the index value. Among all the selected solutions, we compile the one that is chosen the most, at a configurable ratio, and ahead-of-time (AOT) compile them into executables. The rest will be compiled just-in-time (JIT) to balance build cost and runtime overhead.

\paragraph{Online Lightweight Dispatch.}
At runtime, we construct the key from the input arguments of the kernel, perform an $O(1)$ lookup in the index, and find or compile a valid kernel for execution. When CUDA graph is enabled and with proper warmup, the overhead is negligible (see \autoref{sec:apply_overhead}).

\section{Dataset Overview and Evaluation}
\label{sec:dataset}

\subsection{Dataset Overview}

Our dataset includes eight representative kernel types used in LLM inference: GEMM, Ragged and Paged GQA, Ragged and Paged MLA, Fused MoE, RMS Normalization, and Sampling. These cover the core components of modern LLMs, including fused, non-deterministic, and low-bit kernels (e.g., FP8 GEMM). For each kernel type, when certain dimensions (e.g., hidden dimension) are fixed constants, we treat every unique combination of these parameters as a separate definition, resulting in 41 distinct kernel configurations. For workloads, we collected real input instances on SGLang using input traces and curated a dataset of 1,600 workloads through shape-based deduplication and filtering, covering both short- and long-sequence cases. Solutions were generated under the agent framework (Section 4.2) using frontier models, including Gemini 2.5 Pro, Claude Opus 4.1, GPT-5, and OpenAI o3, in CUDA and Triton. Notably, \fib also supports additional DSLs such as CUTLASS and CuTe DSL. In total, 240 solutions were evaluated across all workloads, producing 9,600 evaluation results.

\subsection{Agent Evaluation Settings}

\paragraph{Setup.} To ensure fair and accurate evaluation across multiple models, we design a feedback-loop agent, as shown in \autoref{alg:iterative}. In each iteration, the agent generates a kernel, evaluates it using \fib, and refines the design based on the results until reaching the improvement limit. The best kernel produced across all iterations is selected as the final solution for evaluation. Kernel performance is measured on an NVIDIA B200 GPU.

\begin{algorithm}
\caption{Feedback-loop Agent}
\label{alg:iterative}
\small
\DontPrintSemicolon
\SetKwInOut{KwIn}{Input}
\SetKwInOut{KwOut}{Output}
\KwIn{Definition, Language, Hardware}
\KwOut{Sol$^*$ (best solution)}

$\mathcal{S} \leftarrow \emptyset$\;

$\text{Agent} \leftarrow \text{CodeAgent.Initialize}(\text{Definition}, \text{Language}, \text{Hardware})$\;

$\text{Sol}_0 \leftarrow \text{Agent.Generate}()$\;

\For{$i \leftarrow 0$ \KwTo $N-1$}{
  $\text{Trace}_i \leftarrow \text{FlashInfer-Bench.Benchmark}(\text{Definition}, \text{Sol}_i)$\;

  \If{$\text{Trace}_i.\text{Status} = \textsc{Passed}$}{
    $\mathcal{S} \leftarrow \mathcal{S} \cup \{(\text{Sol}_i, \text{Trace}_i)\}$\;
  }
  $\text{Sol}_{i+1} \leftarrow \text{Agent.Optimize}(\text{Trace}_i)$\;
}

$\text{Sol}^* \leftarrow \arg\max_{(\text{Sol}_i, \text{Trace}_i) \in \mathcal{S}} \text{Trace}_i.\text{Speedup}$\;

\Return{$\text{Sol}^*$}
\end{algorithm}

\paragraph{Metrics.} We adopt the $\text{fast}_p$ metric from KernelBench~\cite{ouyang2025kernelbench}. For a solution, $\text{fast}_p$ represents the proportion of workloads on which a kernel runs more than $p$ times faster than the baseline kernel. Formally,
\begin{equation}
\text{fast}_p = \frac{1}{N} \sum_{i=1}^{N} \mathbf{1}(\text{correct}_i \land \{\text{speedup}_i > p\})
\end{equation}

By varying $p$, we obtain different $\text{fast}_p$ values, forming a curve whose area under the curve (AUC) represents the overall performance of the kernel. This curve captures both kernel correctness and performance. We choose the state-of-the-art kernel library FlashInfer \cite{ye2025flashinfer} as the comparison baseline. If a corresponding FlashInfer kernel is unavailable (e.g., for bf16 GEMM), we use PyTorch instead. For each model, we further average the $\text{fast}_p$ values across all generated solutions to obtain the model's $\text{fast}_p$ curve. Notably, when $p = 0$, the $\text{fast}_p$ value represents the correctness rate of the kernels generated by the agent.

\subsection{Analysis of Agent Capability}

\begin{figure*}
\centering
\includegraphics[width=0.9\linewidth]{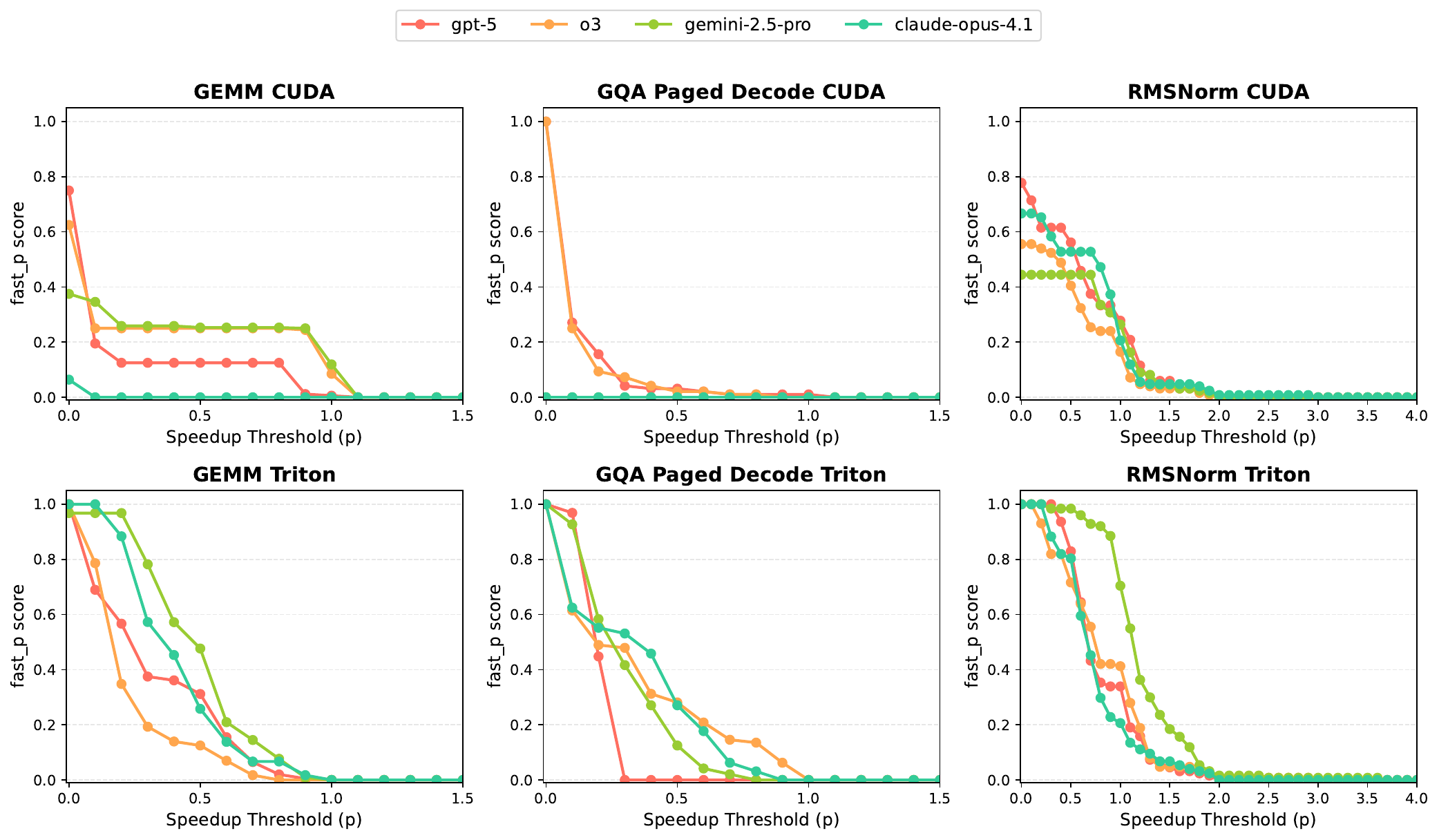}
\caption{The \fastp metric captures both axes of correctness and performance, defined as the fraction of kernel-workload evaluations that are both correct and achieve a speedup over baseline greater than threshold $p$.}
\label{fig:fast_p_grid}
\end{figure*}

The evaluation results are shown in \autoref{fig:fast_p_grid}. We present $\text{fast}_p$ curves for three representative kernels: GEMM, GQA Paged Decoding, and RMSNorm. Results show that LLM performance lags behind humans in most cases: for GEMM (Triton), GQA (CUDA), and GQA (Triton), it achieves less than 50\% of SOTA performance on more than half of the workloads. RMSNorm performs close to or above human level since it is memory-bound, reaching speed limits once memory bandwidth saturates, making it easier to optimize. In contrast, GEMM and GQA are compute-bound, requiring techniques like pipelining and tiling to avoid stalls, so harder to optimize. The strong GEMM (CUDA) performance comes from the agent learning to use library kernels, which will be discussed later.

\paragraph{Most correctness errors come from compilation failures.}
Among all 32 correctness errors, 30 are due to compilation errors, while only 2 are runtime or numerical errors. The numerical errors mainly arise from incorrect padding calculations in dispatched kernels for large input shapes.  We categorize the compilation errors into several representative types:

\begin{enumerate}
  \item \textbf{API Usage Error.} The model may call the correct API but use it incorrectly, which is especially common in \texttt{Triton} generation. For example, the model may attempt to index a \texttt{constexpr} variable, an operation not supported in Triton. We attribute this to the limited amount of Triton-related data during training, which prevents the model from learning correct usage patterns. Occasionally, the model also generates non-existent APIs, such as \texttt{\_\_nv\_bfloat162\_to\_float2}, though these issues are often resolved after several refinement rounds.
  \item \textbf{Host–Device Confusion.} The model sometimes confuses host code with device code. For instance, in a \texttt{Triton} kernel, it may call \texttt{math.log} on the host side instead of \texttt{tl.log}. This likely stems from the model’s limited ability to accurately recognize its current execution context.
  \item \textbf{Datatype Error and Shape Error.} The model sometimes generates code with incorrect datatype usage or mismatched tensor shapes. For example, it may use a \texttt{float} input for the \texttt{\_\_reduce\_add\_sync} function, which only allows integer inputs.
\end{enumerate}

\paragraph{LLMs struggles to correctly apply hardware intrinsics and optimizations.} Although we provided hardware information, including specifications and relevant documentation, the model failed to leverage this knowledge to maximize kernel performance, especially on the latest hardware. For example, in the \texttt{GEMM CUDA} task, models relied solely on \texttt{wmma} for optimization but did not correctly utilize more efficient \texttt{mma} or the new tensor instruction \texttt{tcgen05} available on Blackwell GPUs, leading to suboptimal performance. In the attention task, the model attempted to implement the FlashAttention algorithm in CUDA but only reproduced the online softmax component, without proper use of tiling or tensor cores. These observations suggest that the model's ability to specialize for specific hardware remains limited. Future work could employ reinforcement learning or similar methods to encourage better exploration of hardware capabilities.

\paragraph{GPU programming languages bring trade-off for agent design.}

\begin{figure}[h]
\centering
\includegraphics[width=0.9\linewidth]{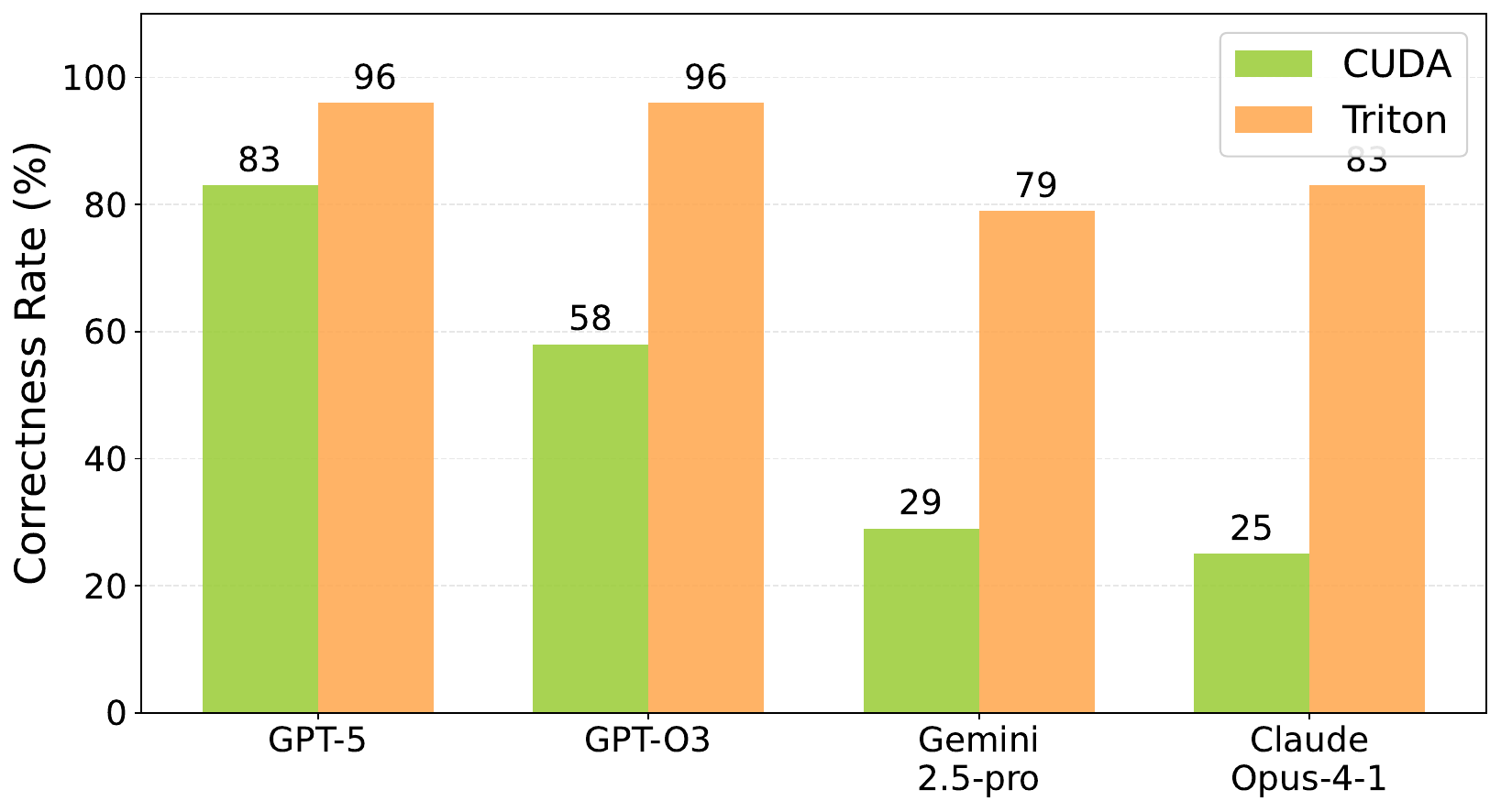}
\caption{Overall correctness of LLM Agents across languages.}
\label{fig:model_correctness_rate}
\end{figure}

GPU programming languages can be roughly divided into two categories: \textbf{high-level abstractions} such as \texttt{Triton} and \texttt{TVM}, which encapsulate low-level hardware details and require only high-level computational logic, and \textbf{low-level languages} such as \texttt{CUDA} and \texttt{PTX}, which demand explicit handling of hardware-specific details. We observe that the model achieves significantly higher correctness (\autoref{fig:model_correctness_rate}) and speed when writing \texttt{Triton} kernels compared to \texttt{CUDA}. We attribute this to several factors. First, \texttt{Triton} code is shorter and less detailed, reducing the demand on the model’s long-context reasoning ability. Second, \texttt{Triton} only requires expressing high-level (tile-level) computation logic, while the compiler automatically determines optimal low-level implementations, enabling the use of advanced hardware features such as \texttt{tcgen05}, which is difficult for agents to exploit in \texttt{CUDA}. However, since \texttt{CUDA} exposes more precise hardware abstractions—such as explicit control over shared memory—it provides a higher performance ceiling. This indicates that improving agents on \texttt{CUDA} offers greater potential for further performance gains.

\paragraph{Agents have learned to call libraries instead of writing raw kernels.} For example, in the \texttt{GEMM CUDA} task, we observed that agents based on \texttt{Gemini-2.5-pro} and \texttt{o3} learned to invoke the \texttt{cuBLAS} library's \texttt{matmul} kernel in some solutions, achieving performance comparable to or exceeding the baseline. This finding has two implications: first, the model demonstrates the ability to leverage its environment to produce near-human-level CUDA code; second, it suggests that the model relies on library calls rather than truly understanding GPU programming, which could become a bottleneck during training. We recommend restricting library access during training for better skill acquisition, while allowing unrestricted use during inference to maximize kernel performance.

\subsection{Agent Generated Kernel Case Study}
\paragraph{GEMM - Compilers help}
For this case study, we analyze GPT-5's generated Triton and CUDA kernels for definition \texttt{gemm\_n4096\_k4096}. The Triton kernel achieves a 4.5$\times$ speedup over the CUDA kernel, with a mean execution time of 0.11ms against 0.5ms on tested workloads.
Full implementation is provided in the Appendix; see \autoref{appendix:triton_gemm} and \autoref{appendix:cuda_gemm}.

The CUDA kernel implements a three-stage tiling strategy with fixed block tiles of 128$\times$256$\times$64 and 8 warps per block. Each warp computes a 64$\times$64 output region using 16 WMMA operations arranged in a 4$\times$4 grid. The implementation manually manages shared memory with skewed padding to avoid bank conflicts and uses vectorized 128-bit loads to maximize memory bandwidth. Overall, the CUDA kernel follows a simple load-sync-compute-sync pattern without inter-tile prefetching or software pipelining, relying primarily on hardware-level intra-tile overlap for compute-memory concurrency.

The Triton kernel implements a similar high-level tiling strategy but leverages Triton's \texttt{autotune} decorator to evaluate four different tile configurations at runtime and select the optimal one based on the workload's M dimension. The kernel benefits from the compiler's built-in software pipelining (configured to depth 4) that overlaps computation with memory prefetching across K-dimension iterations.

The largest performance difference stems from the choice of tensor core primitives. The CUDA kernel uses WMMA instructions, while the Triton kernel automatically targets tcgen05 instructions through its \texttt{tl.dot()} operator. This highlights a critical limitation of CUDA code generation: when new architectures introduce novel intrinsics, these instructions initially lack sufficient training examples in the agent's corpus, causing agents to default to older, suboptimal patterns. In addition, Triton's high-level load/store abstractions (\texttt{tl.load()}, \texttt{tl.store()}) alleviate the need for the agent to manage shared memory allocation, bank conflict avoidance, and memory coalescing. This demonstrates that DSLs with appropriate abstractions enable agents to more effectively explore and apply advanced optimization techniques by reducing the cognitive complexity of coordinating multiple optimization strategies.

\paragraph{GQA Paged Decode: Optimization is Difficult}
The CUDA implementation of GQA paged decode by GPT-5 contains few optimizations, only using online softmax for memory efficiency. Each block processes one batch element and KV head, with scalar FP32 arithmetic for attention computation and basic memory access patterns. It lacks general optimization techniques used among state-of-the-art kernels, such as block-wise tiling to minimize HBM memory traffic, asynchronous execution, and pipelining to overlap GEMM and softmax operations~\cite{shah2024flashattention3}. Full implementation is provided in the \autoref{appendix:cuda_attention}.

We tried explicitly prompting the LLM agent (GPT-5-2025-08-07 on high reasoning mode) with these attention kernel optimization strategies and instructions for CUDA intrinsics, but the LLM agent failed to generate correct kernels that utilize these optimizations in 10 attempts.

This case study shows that although CUDA exposes fine-grained hardware control and enables sophisticated optimizations, pretrained LLMs struggle to leverage these capabilities due to the increased implementation complexity. The agent can recognize high-level optimization strategies when prompted, but cannot correctly coordinate the low-level details required for CUDA implementation. Effectively utilizing CUDA's expressiveness for complex kernels likely requires additional training approaches such as reinforcement learning with high-quality execution feedback or curated datasets of expert kernel implementations.

\paragraph{Summary}
Overall, the case studies reveal that while LLMs like GPT-5 can generate functionally correct kernels, they struggle to match expert-level performance in low-level GPU optimization. High-level abstraction DSLs allow them to achieve competitive efficiency through compiler support, but direct CUDA generation exposes weaknesses in memory management, tiling, and hardware-specific tuning.

\subsection{Kernel Substitution in End-to-end Systems}\label{sec:apply_overhead}

We aim to demonstrate the effectiveness and efficiency of our dynamic kernel substitution mechanism (\fibapplytt). To do this, we provide multiple implementations of the same kernel definition with different speeds, all generated by the agent and validated by \fib, and dynamically substitute them into the serving engine. We expect that faster kernels will lead to lower end-to-end request latency, which would validate the effectiveness of kernel substitution. We also extract the original kernel from the serving engine, restrict the dynamic dispatcher so that it can only dispatch to this kernel, and compare the end-to-end latency with and without substitution enabled, allowing us to isolate the overhead of the substitution mechanism itself.

\begin{figure}[h]
\centering
\includegraphics[width=0.9\linewidth]{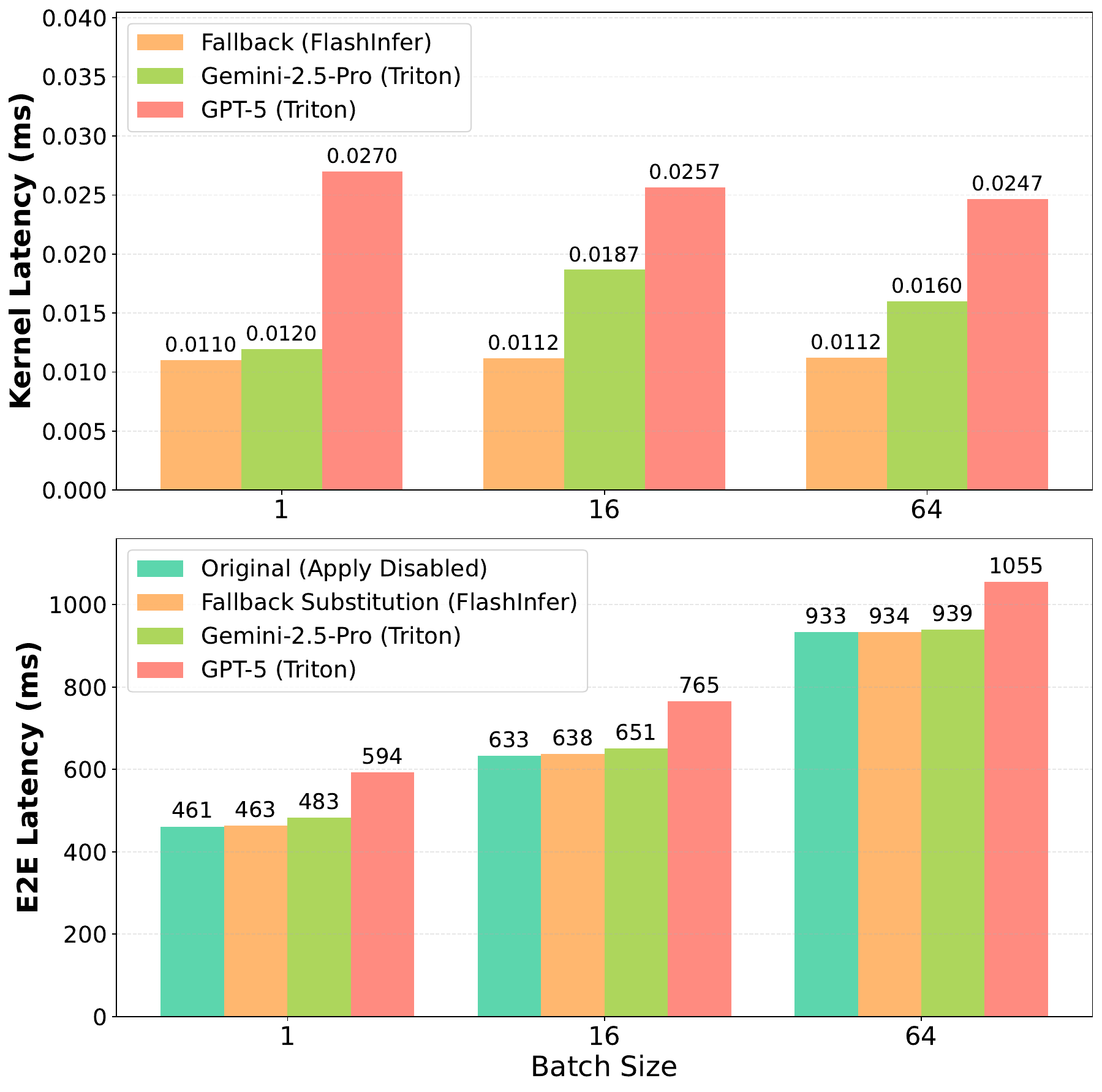}
\caption{Kernel latency and end-to-end latency comparison. (Top) \texttt{fused\_add\_rmsnorm\_h4096} for three implementations across batch sizes 1, 16, and 64. (Bottom) End-to-end request latency comparing the original baseline with different kernel substitution mechanisms. All measurements in milliseconds; lower is better.}
\label{fig:system_evaluation}
\vspace{-10pt}
\end{figure}

\paragraph{Experimental Setup}
We evaluate our system using the Fused Add RMSNorm kernel (hidden size 4096) as a representative case, deployed in SGLang serving Llama-3.1-8B-Instruct. For each configuration, we measure (i) isolated kernel latency from our benchmark traces (Figure~\ref{fig:system_evaluation}, top) and (ii) end-to-end request latency across varying batch sizes/concurrency levels (1, 16, 64). We compare four configurations: \textbf{Original}, native SGLang with the FlashInfer backend and \texttt{apply()} enabled; \textbf{Fallback Substitution}, the baseline FlashInfer implementation but substituted by \texttt{apply()}; \textbf{Gemini-2.5-Pro (Triton)}, a generated kernel faster than the baseline; and \textbf{GPT-5 (Triton)}, a generated kernel slower than the baseline. Each experiment runs with a system warm-up followed by measuring the mean latency over 4 requests with the same input and output length to ensure stable readings.

\paragraph{\texttt{apply()} introduces minimal overhead.}
Our first experiment isolates the fixed cost of \texttt{apply()} by substituting the native kernel with an identical implementation (Fallback Substitution). Profiling shows \texttt{apply()} introduces 1-2 us overhead for each kernel calling. Comparing the \textit{Original} and \textit{Fallback Substitution} configurations in Figure~\ref{fig:system_evaluation} (bottom), we observe that end-to-end overhead is less than 0.8\% across all batch sizes.

\paragraph{\texttt{apply()} translates kernel gains into end-to-end latency improvements.}
Having established minimal overhead, we next show that kernel-level improvements translate into measurable end-to-end gains. The top plot in Figure~\ref{fig:system_evaluation} reports isolated kernel performance: FlashInfer achieves 0.0112 ms at batch size 64, a speedup over Gemini-2.5-Pro Triton kernel (0.0160 ms), while GPT-5 records 0.0247 ms. These benchmark gains carry through to the full system: in the bottom plot, the end-to-end times for FlashInfer Fallback substitution, Gemini-2.5-pro substitution, and GPT-5 substitution with \texttt{apply()} are 934 ms, 939 ms, and 1055 ms, respectively, consistent with the performance of their kernels. This indicates that the efficiency of the kernel is directly translated to the end-to-end efficiency. When we substitute a better kernel, we can achieve better end-to-end efficiency.
 
\section{Related Work}
\label{sec:related_work}
\subsection{LLM for CUDA Generation}

LLMs have recently shown remarkable capabilities in generating GPU kernels, with benchmarks such as KernelBench~\cite{ouyang2025kernelbench} and TritonBench~\cite{li2025tritonbench} systematically evaluating these abilities. These benchmarks are focusing primarily on the evaluation of model capabilities. BackendBench~\cite{saroufim2025backendbench} studied how to use LLMs to generate kernels in PyTorch and integrate the kernel into PyTorch. \fib, in contrast, focuses on the major workload in LLM systems and provides an end-to-end production system that integrates evaluation, validation, and deployment into a unified framework.

Recent advances have explored training LLMs and designing agents to generate efficient kernels. Kevin~\cite{baronio2025kevin} and KernelLLM~\cite{kernelllm2025} developed post-trained models for kernel generation. \citet{dong2025qimengxpiler} designed an agent for the kernel transpiling task, while \citet{wei2025astramultiagentgpukernel} studied agent designs for LLM-based CUDA kernel generation. These agent and model designs are complementary to \fib, which can further evaluate the performance of these agents and models on kernel tasks from real-world LLM systems.

\subsection{Machine Learning for Systems Optimization}

Machine learning has long been applied to compiler and systems optimization. TVM~\cite{chen2018tvm} pioneered automatic tensor program optimization, followed by AutoTVM and Ansor~\cite{zheng2020ansor}, which introduced search-based tuning with learned cost models. More recently, Meta-Schedule~\cite{tvm_metaschedule} generalized these methods via design-space abstraction and policy learning. These systems operate within a fixed optimization space defined by human-crafted templates.

\fib adopts a \emph{generation-based} approach: LLMs directly propose candidate kernels (potentially outside any predefined schedule space), which are then subjected to strict functional validation and performance benchmarking. This expands the effective frontier from searching within a fixed template set to synthesizing new implementation patterns.

\subsection{Kernel libraries and Custom Kernel DSLs}

Highly optimized kernel libraries, such as CUTLASS \citep{cutlass}, cuBLAS \cite{nvidia_cublas_13}, and FlashInfer \citep{ye2025flashinfer}, provide strong baselines for kernels in LLM systems. Domain-specific languages such as Triton \citep{triton} lower the barrier to writing custom GPU kernels. \fib treats these ecosystems as implementation targets for agents, and focuses on connecting candidate kernels to production systems.

Several works specialize in inference workloads. FlashAttention introduced IO-aware exact attention that significantly reduces memory traffic and improves throughput \citep{dao2022flashattention}. Multi-Query Attention reduces the KV cache footprint and improves decoding latency by sharing keys/values across heads \citep{shazeer2019mqa}. Serving engines adopt paged KV caches and scheduling policies (e.g., PagedAttention) to maintain high utilization under ragged, multi-tenant workloads \citep{kwon2023vllm}. \fib captures these realities by grounding tasks in real traces (shape distributions, cache layouts) and by evaluating numerical and batching properties that are critical for correct deployment.

\subsection{LLM Inference Systems}

Frameworks such as vLLM~\cite{kwon2023vllm}, SGLang~\cite{NEURIPS2024_SGLang}, TensorRT-LLM~\cite{tensorrt-llm}, and MLC-LLM~\cite{mlc-llm} demonstrate scalable inference infrastructure for large models. These systems guide \fib's kernel selection--prioritizing modern LLM operations like attention and MoE over legacy operations like convolution—and provide reference implementations. In turn, \fib enables these frameworks to rapidly evaluate and deploy optimized kernels into production, creating a mutually beneficial ecosystem.

\section{Conclusion}
\label{sec:conclusion}

We presented \fib, a systematic approach that closes the loop from AI kernel generation to production impact. At its core, the \fitrace schema standardizes operator contracts, real serving workloads, candidate implementations, and immutable evaluations. Built atop this, our benchmark measures deterministic, low-precision, and sampling kernels and, via \texttt{apply()}, can substitute the best validated kernel into engines such as SGLang and vLLM with zero code changes.

Complementing the framework, our live leaderboard continuously tracks frontier models’ GPU programming capabilities over real-world and LLM workloads. Our evaluation led to three practical takeaways: (1) compilation is the dominant failure mode, (2) models struggle to exploit hardware features, and (3) Language choice is a trade-off--Triton yields high correctness and usability, while CUDA, when successful, reaches higher peak performance. End-to-end, dynamic substitution adds negligible overhead and reliably converts kernel-level gains into lower latency and higher throughput in LLM serving.

Limitations and next steps: Our current scope does not yet cover multi-GPU or communication kernels, and the range of supported models, hardware devices, and programming languages remains limited. Future works can further extend the \fitrace dataset breadth, improve kernel correctness verification to prevent reward hacking and ensure reliable benchmarking outcomes, and develop kernel agents and fine-tuned models for LLM systems based on the \fib feedback loop.

\section*{Acknowledgements}

This work is supported in part by Bosch and gifts from NVIDIA and Google, and we also acknowledge the support of DGX B200 from NVIDIA. We would also like to thank, listed alphabetically, Databricks, the FlashInfer team, the GPUMODE team, the HuggingFace team, the SGLang team, the TensorRT-LLM team, the vLLM team, and xAI, as well as Zhuoming Chen, Weihua Du, Bohan Hou, Hongyi Jin, Ruihang Lai, Mark Saroufim, Xinyu Yang, Yilong Zhao, and Haizhong Zheng for their insightful feedback.

% TODO: Add acknowledgements (for camera-ready version only)

% Bibliography
\bibliography{reference}

\begin{thebibliography}{21}
\providecommand{\natexlab}[1]{#1}
\providecommand{\url}[1]{\texttt{#1}}
\expandafter\ifx\csname urlstyle\endcsname\relax
  \providecommand{\doi}[1]{doi: #1}\else
  \providecommand{\doi}{doi: \begingroup \urlstyle{rm}\Url}\fi

\bibitem[Baronio et~al.(2025)Baronio, Marsella, Pan, Guo, and Alberti]{baronio2025kevin}
Baronio, C., Marsella, P., Pan, B., Guo, S., and Alberti, S.
\newblock Kevin: Multi-turn rl for generating {CUDA} kernels, 2025.
\newblock URL \url{https://arxiv.org/abs/2507.11948}.

\bibitem[Chen et~al.(2018)Chen, Moreau, Jiang, Zheng, Yan, Cowan, Shen, Wang, Hu, Ceze, Guestrin, and Krishnamurthy]{chen2018tvm}
Chen, T., Moreau, T., Jiang, Z., Zheng, L., Yan, E., Cowan, M., Shen, H., Wang, L., Hu, Y., Ceze, L., Guestrin, C., and Krishnamurthy, A.
\newblock Tvm: An automated end-to-end optimizing compiler for deep learning.
\newblock In \emph{Proceedings of the 13th USENIX Symposium on Operating Systems Design and Implementation (OSDI)}, Carlsbad, CA, USA, 2018. USENIX Association.

\bibitem[Dao et~al.(2022)Dao, Fu, Ermon, Rudra, and R\'{e}]{dao2022flashattention}
Dao, T., Fu, D.~Y., Ermon, S., Rudra, A., and R\'{e}, C.
\newblock Flashattention: fast and memory-efficient exact attention with io-awareness.
\newblock In \emph{Proceedings of the 36th International Conference on Neural Information Processing Systems}, NIPS '22, Red Hook, NY, USA, 2022. Curran Associates Inc.
\newblock ISBN 9781713871088.

\bibitem[Dong et~al.(2025)Dong, Wen, Bi, Huang, Guo, Xu, Xu, Song, Hao, Zhou, Chen, Guo, and Chen]{dong2025qimengxpiler}
Dong, S., Wen, Y., Bi, J., Huang, D., Guo, J., Xu, J., Xu, R., Song, X., Hao, Y., Zhou, X., Chen, T., Guo, Q., and Chen, Y.
\newblock Qimeng-xpiler: Transcompiling tensor programs for deep learning systems with a neural-symbolic approach, 2025.
\newblock URL \url{https://arxiv.org/abs/2505.02146}.
\newblock Accepted to OSDI 2025.

\bibitem[Fisches et~al.(2025)Fisches, Paliskara, Guo, Zhang, Spisak, Cummins, Leather, Synnaeve, Isaacson, Markosyan, and Saroufim]{kernelllm2025}
Fisches, Z.~V., Paliskara, S., Guo, S., Zhang, A., Spisak, J., Cummins, C., Leather, H., Synnaeve, G., Isaacson, J., Markosyan, A., and Saroufim, M.
\newblock Kernelllm: Making kernel development more accessible, 6 2025.
\newblock URL \url{https://huggingface.co/facebook/KernelLLM}.
\newblock Corresponding authors: Aram Markosyan, Mark Saroufim.

\bibitem[Kwon et~al.(2023)Kwon, Li, Zhuang, Sheng, Zheng, Yu, Gonzalez, Zhang, and Stoica]{kwon2023vllm}
Kwon, W., Li, Z., Zhuang, S., Sheng, Y., Zheng, L., Yu, C.~H., Gonzalez, J., Zhang, H., and Stoica, I.
\newblock Efficient memory management for large language model serving with pagedattention.
\newblock In \emph{Proceedings of the 29th Symposium on Operating Systems Principles}, SOSP '23, pp.\  611–626, New York, NY, USA, 2023. Association for Computing Machinery.
\newblock ISBN 9798400702297.
\newblock \doi{10.1145/3600006.3613165}.
\newblock URL \url{https://doi.org/10.1145/3600006.3613165}.

\bibitem[Li et~al.(2025)Li, Li, Gao, Shi, Li, Wang, Huang, Wang, Wang, Han, Liu, and Sun]{li2025tritonbench}
Li, J., Li, S., Gao, Z., Shi, Q., Li, Y., Wang, Z., Huang, J., Wang, H., Wang, J., Han, X., Liu, Z., and Sun, M.
\newblock Tritonbench: Benchmarking large language model capabilities for generating triton operators, 2025.
\newblock URL \url{https://arxiv.org/abs/2502.14752}.

\bibitem[{MLC team}(2023-2025)]{mlc-llm}
{MLC team}.
\newblock {MLC-LLM}, 2023-2025.
\newblock URL \url{https://github.com/mlc-ai/mlc-llm}.

\bibitem[NVI(2025)]{nvidia_cublas_13}
\emph{cuBLAS Library}.
\newblock NVIDIA, October 2025.
\newblock URL \url{https://docs.nvidia.com/cuda/pdf/CUBLAS_Library.pdf}.
\newblock CUDA Toolkit Documentation.

\bibitem[{NVIDIA}(2025)]{tensorrt-llm}
{NVIDIA}.
\newblock Tensorrt llm, 2025.
\newblock URL \url{https://github.com/NVIDIA/TensorRT-LLM}.
\newblock GitHub repository.

\bibitem[Ouyang et~al.(2025)Ouyang, Guo, Arora, Zhang, Hu, R{\'e}, and Mirhoseini]{ouyang2025kernelbench}
Ouyang, A., Guo, S., Arora, S., Zhang, A.~L., Hu, W., R{\'e}, C., and Mirhoseini, A.
\newblock Kernelbench: Can {LLMs} write efficient {GPU} kernels?, 2025.
\newblock URL \url{https://arxiv.org/abs/2502.10517}.

\bibitem[Saroufim et~al.(2025)Saroufim, Wang, Maher, Paliskara, Wang, Sefati, and Candales]{saroufim2025backendbench}
Saroufim, M., Wang, J., Maher, B., Paliskara, S., Wang, L., Sefati, S., and Candales, M.
\newblock Backendbench: An evaluation suite for testing how well llms and humans can write pytorch backends, 2025.
\newblock URL \url{https://github.com/meta-pytorch/BackendBench}.

\bibitem[Shah et~al.(2024)Shah, Bikshandi, Zhang, Thakkar, Ramani, and Dao]{shah2024flashattention3}
Shah, J., Bikshandi, G., Zhang, Y., Thakkar, V., Ramani, P., and Dao, T.
\newblock Flashattention-3: Fast and accurate attention with asynchrony and low-precision.
\newblock In \emph{Proceedings of the 38th International Conference on Neural Information Processing Systems}, NIPS '24, Red Hook, NY, USA, 2024. Curran Associates Inc.

\bibitem[Shao et~al.(2022)Shao, Zhou, Feng, Hou, Lai, Jin, Lin, Masuda, Yu, and Chen]{tvm_metaschedule}
Shao, J., Zhou, X., Feng, S., Hou, B., Lai, R., Jin, H., Lin, W., Masuda, M., Yu, C.~H., and Chen, T.
\newblock Tensor program optimization with probabilistic programs, 2022.
\newblock URL \url{https://arxiv.org/abs/2205.13603}.

\bibitem[Shazeer(2019)]{shazeer2019mqa}
Shazeer, N.
\newblock Fast transformer decoding: One write-head is all you need, 2019.
\newblock URL \url{https://arxiv.org/abs/1911.02150}.

\bibitem[Thakkar et~al.(2023)Thakkar, Ramani, Cecka, Shivam, Lu, Yan, Kosaian, Hoemmen, Wu, Kerr, Nicely, Merrill, Blasig, Qiao, Majcher, Springer, Hohnerbach, Wang, and Gupta]{cutlass}
Thakkar, V., Ramani, P., Cecka, C., Shivam, A., Lu, H., Yan, E., Kosaian, J., Hoemmen, M., Wu, H., Kerr, A., Nicely, M., Merrill, D., Blasig, D., Qiao, F., Majcher, P., Springer, P., Hohnerbach, M., Wang, J., and Gupta, M.
\newblock {CUTLASS}, January 2023.
\newblock URL \url{https://github.com/NVIDIA/cutlass}.

\bibitem[Tillet et~al.(2019)Tillet, Kung, and Cox]{triton}
Tillet, P., Kung, H.~T., and Cox, D.
\newblock Triton: an intermediate language and compiler for tiled neural network computations.
\newblock In \emph{Proceedings of the 3rd ACM SIGPLAN International Workshop on Machine Learning and Programming Languages}, MAPL 2019, pp.\  10–19, New York, NY, USA, 2019. Association for Computing Machinery.
\newblock ISBN 9781450367196.
\newblock \doi{10.1145/3315508.3329973}.
\newblock URL \url{https://doi.org/10.1145/3315508.3329973}.

\bibitem[Wei et~al.(2025)Wei, Sun, Seenichamy, Song, Ouyang, Mirhoseini, Wang, and Aiken]{wei2025astramultiagentgpukernel}
Wei, A., Sun, T., Seenichamy, Y., Song, H., Ouyang, A., Mirhoseini, A., Wang, K., and Aiken, A.
\newblock Astra: A multi-agent system for gpu kernel performance optimization, 2025.
\newblock URL \url{https://arxiv.org/abs/2509.07506}.

\bibitem[Ye et~al.(2025)Ye, Chen, Lai, Lin, Zhang, Wang, Chen, Kasikci, Grover, Krishnamurthy, and Ceze]{ye2025flashinfer}
Ye, Z., Chen, L., Lai, R., Lin, W., Zhang, Y., Wang, S., Chen, T., Kasikci, B., Grover, V., Krishnamurthy, A., and Ceze, L.
\newblock Flashinfer: Efficient and customizable attention engine for llm inference serving.
\newblock \emph{arXiv preprint arXiv:2501.01005}, 2025.
\newblock URL \url{https://arxiv.org/abs/2501.01005}.

\bibitem[Zheng et~al.(2020)Zheng, Jia, Sun, Wu, Yu, Haj-Ali, Wang, Yang, Zhuo, Sen, Gonzalez, and Stoica]{zheng2020ansor}
Zheng, L., Jia, C., Sun, M., Wu, Z., Yu, C.~H., Haj-Ali, A., Wang, Y., Yang, J., Zhuo, D., Sen, K., Gonzalez, J.~E., and Stoica, I.
\newblock Ansor: generating high-performance tensor programs for deep learning.
\newblock In \emph{Proceedings of the 14th USENIX Conference on Operating Systems Design and Implementation}, OSDI'20, USA, 2020. USENIX Association.
\newblock ISBN 978-1-939133-19-9.

\bibitem[Zheng et~al.(2025)Zheng, Yin, Xie, Sun, Huang, Yu, Cao, Kozyrakis, Stoica, Gonzalez, Barrett, and Sheng]{NEURIPS2024_SGLang}
Zheng, L., Yin, L., Xie, Z., Sun, C., Huang, J., Yu, C.~H., Cao, S., Kozyrakis, C., Stoica, I., Gonzalez, J.~E., Barrett, C., and Sheng, Y.
\newblock Sglang: efficient execution of structured language model programs.
\newblock In \emph{Proceedings of the 38th International Conference on Neural Information Processing Systems}, NIPS '24, Red Hook, NY, USA, 2025. Curran Associates Inc.
\newblock ISBN 9798331314385.

\end{thebibliography}
\bibliographystyle{mlsys2025}

% Appendix
\appendix
\clearpage
\appendix
\onecolumn
\begin{adjustwidth}{0.8em}{0pt}

\section{FlashInfer Trace Examples}
\label{appendix:trace}
This appendix provides concrete examples of the FlashInfer Trace format.

\subsection{GEMM}
\label{appendix:trace_gemm}
We start with the trace definition for a general matrix multiplication kernel.

\captionsetup{type=listing}

\begin{adjustwidth}{0.5em}{0pt}
\begingroup
  \begin{minted}[fontsize=\small, breaklines]{json}
{
  "name": "gemm_n128_k2048",
  "description": "GEMM C = A @ B.T. Captured from Qwen 3 30B A3B moe.gate.",
  "op_type": "gemm",
  "tags": [
    "status:verified",
    "model:qwen3-30b-a3b"
  ],
  "axes": {
    "M": { "type": "var" },
    "N": { "type": "const", "value": 128 },
    "K": { "type": "const", "value": 2048 }
  },
  "inputs": {
    "A": { "shape": ["M", "K"], "dtype": "float16" },
    "B": { "shape": ["N", "K"], "dtype": "float16" }
  },
  "outputs": {
    "C": { "shape": ["M", "N"], "dtype": "float16" }
  },
  "reference": "import torch\n\ndef run(A, B):\n    C = torch.matmul(A, B.T)\n    return C"
}
\end{minted}
\endgroup
\end{adjustwidth}

A corresponding generated solution object may look as follows:
\begin{adjustwidth}{0.5em}{0pt}
\begingroup
  \begin{minted}[fontsize=\small, breaklines]{json}
{
  "name": "claude-opus-4-1-20250805_triton_a20c42",
  "definition": "gemm_n128_k2048",
  "author": "claude-opus-4-1-20250805",
  "spec": {
    "language": "triton",
    "target_hardware": [
      "B200"
    ],
    "entry_point": "main.py::run",
    "dependencies": []
  },
  "sources": [
    {
      "path": "main.py",
      "content": "<source code omitted>"
    }
  ],
  "description": "claude-opus-4-1-20250805 optimized kernel for gemm_n128_k2048 (round 1)"
}
  \end{minted}
\endgroup
\end{adjustwidth}
\clearpage

A sample workload that instantiates this definition is shown below:

\begin{adjustwidth}{0.5em}{0pt}
\begingroup
  \begin{minted}[fontsize=\small, breaklines]{json}
{
  "definition": "gemm_n128_k2048",
  "solution": null,
  "workload": {
    "uuid": "6ba7c7de-dc5a-48d2-8ada-1382feb5ceac",
    "axes": { "M": 6 },
    "inputs": {
      "A": { "type": "random" },
      "B": { "type": "random" }
    }
  },
  "evaluation": null
}
  \end{minted}
\endgroup
\end{adjustwidth}

Evaluating the solution on this workload on an NVIDIA B200 GPU yields the following trace record:

\begin{adjustwidth}{0.5em}{0pt}
\begingroup
  \begin{minted}[fontsize=\small, breaklines]{json}
{
  "definition": "gemm_n128_k2048",
  "workload": {
    "axes": { "M": 6 },
    "inputs": {
      "A": { "type": "random" },
      "B": { "type": "random" }
    },
    "uuid": "6ba7c7de-dc5a-48d2-8ada-1382feb5ceac"
  },
  "solution": "claude-opus-4-1-20250805_triton_a20c42",
  "evaluation": {
    "status": "PASSED",
    "environment": {
      "hardware": "NVIDIA B200",
      "libs": {
        "torch": "2.8.0+cu128",
        "triton": "3.4.0",
        "cuda": "12.8"
      }
    },
    "timestamp": "2025-10-16T01:10:32.241021",
    "log": "",
    "correctness": {
      "max_relative_error": 0,
      "max_absolute_error": 0,
      "extra": null
    },
    "performance": {
      "latency_ms": 0.023046740692633086,
      "reference_latency_ms": 0.025240250456929125,
      "speedup_factor": 1.0951765715399921
    }
  }
}
  \end{minted}
\endgroup
\end{adjustwidth}

\clearpage

\subsection{Attention}
\label{appendix:trace_attention}
We next present a more complex example based on a paged grouped-query attention decode operator. Compared to the GEMM case, this operator has more complex interfaces and a non-trivial reference implementation.
\begin{adjustwidth}{0.5em}{0pt}
\begingroup
  \begin{minted}[fontsize=\small, breaklines]{json}
{
  "name": "gqa_paged_decode_h32_kv4_d128_ps1",
  "description": "Batched Grouped Query Attention decode with a paged KV cache.",
  "op_type": "gqa_paged",
  "tags": [
    "stage:decode",
    "status:verified",
    "model:qwen3-30b-a3b"
  ],
  "axes": {
    "batch_size": {
      "type": "var",
      "description": "Total number of query tokens."
    },
    "num_qo_heads": {
      "type": "const",
      "value": 32
    },
    "num_kv_heads": {
      "type": "const",
      "value": 4
    },
    "head_dim": {
      "type": "const",
      "value": 128
    },
    "num_pages": {
      "type": "var"
    },
    "page_size": {
      "type": "const",
      "value": 1
    },
    "len_indptr": {
      "type": "var",
      "description": "Length of kv_indptr array."
    },
    "num_kv_indices": {
      "type": "var",
      "description": "Total number of KV page indices."
    }
  },
  "constraints": [
    "len_indptr == batch_size + 1",
    "num_kv_indices == kv_indptr[-1].item()"
  ],
  "inputs": {
    "q": {
      "shape": ["batch_size", "num_qo_heads", "head_dim"],
      "dtype": "bfloat16"
    },
    "k_cache": {
      "shape": ["num_pages", "page_size", "num_kv_heads", "head_dim"],
      "dtype": "bfloat16"
    },
    "v_cache": {
      "shape": ["num_pages", "page_size", "num_kv_heads", "head_dim"],
      "dtype": "bfloat16"
    },
    "kv_indptr": {
      "shape": ["len_indptr"],
      "dtype": "int32",
      "description": "KV page offsets for each sequence."
    },
    "kv_indices": {
      "shape": ["num_kv_indices"],
      "dtype": "int32",
      "description": "Page IDs for KV cache lookups."
    },
    "sm_scale": {
      "shape": null,
      "dtype": "float32",
      "description": "Softmax scale. Default is (1/sqrt(head_dim))."
    }
  },
  "outputs": {
    "output": {
      "shape": ["batch_size", "num_qo_heads", "head_dim"],
      "dtype": "bfloat16"
    },
    "lse": {
      "shape": ["batch_size", "num_qo_heads"],
      "dtype": "float32",
      "description": "The 2-based log-sum-exp of attention logits."
    }
  },
  "reference": "<reference code shown below>"
}
  \end{minted}
\endgroup
\end{adjustwidth}
The corresponding PyTorch reference implementation is shown below:
\captionsetup{type=listing}
\begin{adjustwidth}{0.5em}{0pt}
\begin{minted}{python}
import torch
import math


@torch.no_grad()
def run(q, k_cache, v_cache, kv_indptr, kv_indices, sm_scale):
    batch_size, num_qo_heads, head_dim = q.shape
    _, page_size, num_kv_heads, _ = k_cache.shape
    len_indptr = kv_indptr.shape[0]
    num_kv_indices = kv_indices.shape[0]

    # Check constants
    assert num_qo_heads == 32
    assert num_kv_heads == 4
    assert head_dim == 128
    assert page_size == 1

    # Check constraints
    assert len_indptr == batch_size + 1
    assert num_kv_indices == kv_indptr[-1].item()

    device = q.device

    output = torch.zeros(
        (batch_size, num_qo_heads, head_dim), dtype=torch.bfloat16, device=device
    )
    lse = torch.full(
        (batch_size, num_qo_heads), -float("inf"), dtype=torch.float32, device=device
    )

    gqa_ratio = num_qo_heads // num_kv_heads

    k_cache_flat = k_cache.squeeze(1).to(
        torch.float32
    )  # [num_pages, num_kv_heads, head_dim]
    v_cache_flat = v_cache.squeeze(1).to(
        torch.float32
    )  # [num_pages, num_kv_heads, head_dim]

    for b in range(batch_size):
        page_start = int(kv_indptr[b].item())
        page_end = int(kv_indptr[b + 1].item())

        if page_start >= page_end:
            # No KV cache for this batch element
            output[b].zero_()
            continue

        # Pages are the token indices for page_size=1
        token_indices = kv_indices[page_start:page_end].to(torch.long)
        # Number of tokens is the number of pages for page_size=1
        num_tokens = token_indices.shape[0]

        if num_tokens == 0:
            output[b].zero_()
            continue

        # Get Q, K, V for this batch
        k_batch = k_cache_flat[token_indices]  # [num_tokens, num_kv_heads, head_dim]
        v_batch = v_cache_flat[token_indices]  # [num_tokens, num_kv_heads, head_dim]
        q_batch = q[b].to(torch.float32)  # [num_qo_heads, head_dim]

        for h in range(num_qo_heads):
            # Find corresponding KV head for GQA
            kv_head = h // gqa_ratio

            q_head = q_batch[h]  # [head_dim]
            k_head = k_batch[:, kv_head]  # [num_tokens, head_dim]
            v_head = v_batch[:, kv_head]  # [num_tokens, head_dim]

            logits = torch.matmul(q_head, k_head.T)  # [num_tokens]
            logits_scaled = logits * sm_scale

            # Compute 2-base LSE
            lse[b, h] = torch.logsumexp(logits_scaled, dim=-1) / math.log(2.0)

            attn = torch.softmax(logits_scaled, dim=-1)  # [num_tokens]
            out_head = torch.matmul(attn, v_head)  # [head_dim]
            output[b, h] = out_head.to(torch.bfloat16)

    return output, lse
\end{minted}

\clearpage
\end{adjustwidth}
A potential generated Triton solution can be represented as follows:
\begin{adjustwidth}{0.5em}{0pt}
\begingroup
  \begin{minted}[fontsize=\small, breaklines]{json}
{
  "name": "claude-opus-4-1_triton_de54a2",
  "definition": "gqa_paged_decode_h32_kv4_d128_ps1",
  "description": "claude-opus-4-1-20250805 optimized kernel (round 5)",
  "author": "claude-opus-4-1-20250805",
  "spec": {
    "language": "triton",
    "target_hardware": [
      "B200"
    ],
    "entry_point": "main.py::run",
    "dependencies": []
  },
  "sources": [
    {
      "path": "main.py",
      "content": "<source code omitted>"
    }
  ]
}
  \end{minted}
\endgroup
\end{adjustwidth}
As in the GEMM case, we can capture a concrete workload instance for this definition. In this example, we have a scalar input and some other inputs loaded from a safetensors dump:
\begin{adjustwidth}{0.5em}{0pt}
\begingroup
  \begin{minted}[fontsize=\small, breaklines]{json}
{
  "definition": "gqa_paged_decode_h32_kv4_d128_ps1",
  "solution": null,
  "workload": {
    "uuid": "0c2489b2-f878-428b-b1bd-d0c6d4c39338",
    "axes": {
      "batch_size": 1,
      "num_pages": 8,
      "len_indptr": 2,
      "num_kv_indices": 7
    },
    "inputs": {
      "q": { "type": "random" },
      "k_cache": { "type": "random" },
      "v_cache": { "type": "random" },
      "kv_indptr": {
        "type": "safetensors",
        "path": "/path/to/safetensor",
        "tensor_key": "kv_indptr"
      },
      "kv_indices": {
        "type": "safetensors",
        "path": "/path/to/safetensor",
        "tensor_key": "kv_indices"
      },
      "sm_scale": {
        "type": "scalar",
        "value": 0.0883883461356163
      }
    }
  },
  "evaluation": null
}
  \end{minted}
\endgroup
\end{adjustwidth}
\clearpage

Evaluating the above solution on this workload produces the following trace record:
\begin{adjustwidth}{0.5em}{0pt}
\begingroup
  \begin{minted}[fontsize=\small, breaklines]{json}
{
  "definition": "gqa_paged_decode_h32_kv4_d128_ps1",
  "workload": {
    "axes": {
      "batch_size": 1,
      "num_pages": 8,
      "len_indptr": 2,
      "num_kv_indices": 7
    },
    "inputs": {
      "q": { "type": "random" },
      "k_cache": { "type": "random" },
      "v_cache": { "type": "random" },
      "kv_indptr": {
        "type": "safetensors",
        "path": "/path/to/safetensor",
        "tensor_key": "kv_indptr"
      },
      "kv_indices": {
        "type": "safetensors",
        "path": "/path/to/safetensor",
        "tensor_key": "kv_indices"
      },
      "sm_scale": {
        "type": "scalar",
        "value": 0.0883883461356163
      }
    }
    "uuid": "0c2489b2-f878-428b-b1bd-d0c6d4c39338"
  },
  "solution": "claude-opus-4-1_triton_de54a2",
  "evaluation": {
    "status": "PASSED",
    "environment": {
      "hardware": "NVIDIA B200",
      "libs": {
        "torch": "2.8.0+cu128",
        "triton": "3.4.0",
        "cuda": "12.8"
      }
    },
    "timestamp": "2025-10-16T01:24:16.694452",
    "log": "",
    "correctness": {
      "max_relative_error": 0.01480561401695013,
      "max_absolute_error": 0.00048828125,
      "extra": null
    },
    "performance": {
      "latency_ms": 0.02266162589486805,
      "reference_latency_ms": 29.439284915015815,
      "speedup_factor": 1299.0808802329861
    }
  }
}
  \end{minted}
\endgroup
\end{adjustwidth}

\clearpage

\section{Agent-Generated Kernels}

This appendix contains the full source code for the agent-generated kernels discussed in the case studies. 

\subsection{Triton GEMM Kernel}
\label{appendix:triton_gemm}

\begin{itemize}
  \item \textbf{Language} — Triton.
  \item \textbf{Source} — Generated by GPT-5 from the \texttt{gemm\_n4096\_k4096} definition.
  \item \textbf{Correctness} — Correct across all evaluated workloads.
  \item \textbf{Performance (vs.\ baseline \texttt{torch.nn.functional})} — Speedup is $\mathbf{0.2\times}$ across all workload, with a best case of $\mathbf{0.6\times}$ on one workload.
  \item \textbf{Throughput (TFLOPs)} — For a problem $(M, N{=}4096, K{=}4096)$ with wall-clock time $t$ seconds,
  $$
  \mathrm{TFLOPs} \;=\; \frac{2MNK}{t \cdot 10^{12}}.
  $$
\end{itemize}

\captionsetup{type=listing}
\begin{adjustwidth}{0.5em}{0pt}
\begin{minted}{python}
import math
import torch
import triton
import triton.language as tl


@triton.autotune(
    configs=[
        triton.Config({'BLOCK_M': 128, 'BLOCK_N': 128, 'BLOCK_K': 64}, num_warps=8, num_stages=4),
        triton.Config({'BLOCK_M': 64, 'BLOCK_N': 256, 'BLOCK_K': 64}, num_warps=8, num_stages=4),
        triton.Config({'BLOCK_M': 256, 'BLOCK_N': 64, 'BLOCK_K': 64}, num_warps=8, num_stages=4),
        triton.Config({'BLOCK_M': 128, 'BLOCK_N': 128, 'BLOCK_K': 128}, num_warps=8, num_stages=4),
    ],
    key=['M'],
)
@triton.jit
def _gemm_n4096_k4096_kernel(
    A_ptr, B_ptr, C_ptr,
    M, N, K,
    stride_am, stride_ak,
    stride_bn, stride_bk,
    stride_cm, stride_cn,
    BLOCK_M: tl.constexpr, BLOCK_N: tl.constexpr, BLOCK_K: tl.constexpr,
):
    tl.static_assert(BLOCK_K % 16 == 0, "BLOCK_K must be a multiple of 16 for tensor cores")
    pid_m = tl.program_id(0)
    pid_n = tl.program_id(1)

    offs_m = pid_m * BLOCK_M + tl.arange(0, BLOCK_M)
    offs_n = pid_n * BLOCK_N + tl.arange(0, BLOCK_N)

    acc = tl.zeros((BLOCK_M, BLOCK_N), dtype=tl.float32)

    M_mask = offs_m[:, None] < M
    N_mask = offs_n[None, :] < N

    for k0 in range(0, K, BLOCK_K):
        offs_k = k0 + tl.arange(0, BLOCK_K)

        # Pointers
        a_ptrs = A_ptr + (offs_m[:, None] * stride_am + offs_k[None, :] * stride_ak)
        b_ptrs = B_ptr + (offs_n[:, None] * stride_bn + offs_k[None, :] * stride_bk)

        a = tl.load(a_ptrs, mask=M_mask & (offs_k[None, :] < K), other=0.0)
        b = tl.load(b_ptrs, mask=(offs_n[:, None] < N) & (offs_k[None, :] < K), other=0.0)

        acc += tl.dot(a, tl.trans(b))

    c_ptrs = C_ptr + (offs_m[:, None] * stride_cm + offs_n[None, :] * stride_cn)
    tl.store(c_ptrs, acc.to(tl.float16), mask=M_mask & N_mask)


def run(A, B, **kwargs):
    """
    Compute C = A @ B.T where:
    - A: [M, 4096] float16
    - B: [4096, 4096] float16
    Returns C: [M, 4096] float16
    """
    if not isinstance(A, torch.Tensor) or not isinstance(B, torch.Tensor):
        raise TypeError("A and B must be torch.Tensor")

    if A.ndim != 2 or B.ndim != 2:
        raise ValueError(f"Expected 2D tensors, got A.ndim={A.ndim}, B.ndim={B.ndim}")

    M, K_a = A.shape
    N_b, K_b = B.shape

    if K_a != 4096 or K_b != 4096 or N_b != 4096:
        raise ValueError(f"Invalid shapes: A is {A.shape}, B is {B.shape}; expected A: [M,4096], B: [4096,4096]")

    # Dtype checks/conversions
    if A.dtype != torch.float16:
        A = A.to(torch.float16)
    if B.dtype != torch.float16:
        B = B.to(torch.float16)

    # Device management
    orig_dev_A = A.device
    orig_dev_B = B.device
    any_cuda_input = (A.is_cuda or B.is_cuda)

    if not torch.cuda.is_available():
        if any_cuda_input:
            raise RuntimeError("CUDA is not available but one or more inputs are CUDA tensors.")
        # Triton requires CUDA; no CPU fallback provided
        raise RuntimeError("No CUDA device is available.")

    # Choose compute device
    if A.is_cuda:
        compute_device = A.device
    elif B.is_cuda:
        compute_device = B.device
    else:
        compute_device = torch.device(f"cuda:{torch.cuda.current_device()}")

    # Move to compute device and make contiguous for optimal access
    A_dev = A.to(device=compute_device, non_blocking=True).contiguous()
    B_dev = B.to(device=compute_device, non_blocking=True).contiguous()

    # Allocate output on compute device
    N = 4096
    K = 4096
    C_dev = torch.empty((M, N), dtype=torch.float16, device=compute_device)

    # Kernel launch parameters
    def grid(meta):
        return (triton.cdiv(M, meta['BLOCK_M']), triton.cdiv(N, meta['BLOCK_N']))

    # Call kernel
    _gemm_n4096_k4096_kernel[grid](
        A_dev, B_dev, C_dev,
        M, N, K,
        A_dev.stride(0), A_dev.stride(1),
        B_dev.stride(0), B_dev.stride(1),
        C_dev.stride(0), C_dev.stride(1),
    )

    # Decide output device: preserve original locations; if both were CPU, return CPU; otherwise prefer A's device if CUDA, else B's
    if orig_dev_A.type == 'cpu' and orig_dev_B.type == 'cpu':
        out_device = torch.device('cpu')
    elif orig_dev_A.type == 'cuda':
        out_device = orig_dev_A
    elif orig_dev_B.type == 'cuda':
        out_device = orig_dev_B
    else:
        out_device = torch.device('cpu')

    C_out = C_dev if C_dev.device == out_device else C_dev.to(out_device, non_blocking=True)
    return C_out
\end{minted}
\end{adjustwidth}

\clearpage

\subsection{CUDA GEMM Kernel}
\label{appendix:cuda_gemm}

\begin{itemize}
  \item \textbf{Language} — CUDA.
  \item \textbf{Source} — Auto-generated by Gemini-2.5-Pro for the \texttt{gemm\_n4096\_k4096} definition.
  \item \textbf{Implementation method} — Invokes \texttt{cuBLAS} \texttt{matmul}.
  \item \textbf{Correctness} — Correct across all evaluated workloads.
  \item \textbf{Performance (vs.\ baseline \texttt{torch.nn.functional})} — Speedup is $\mathbf{0.97\times}$ across all workload, with a best case of $\mathbf{1.03\times}$ on one workload.
\end{itemize}

\texttt{kernel.h}:\\

\captionsetup{type=listing}
\begin{adjustwidth}{0.5em}{0pt}
\begin{minted}{cpp}
#pragma once

// Include guard
#ifndef GEMM_N4096_K4096_KERNEL_H
#define GEMM_N4096_K4096_KERNEL_H

#include <cuda_fp16.h>
#include <cuda_runtime.h>

/**
 * @brief Host-side launcher for the GEMM operation C = A * B.T.
 *
 * This function launches a cuBLAS-based GEMM kernel optimized for
 * B200-class GPUs with N=4096 and K=4096. It computes C = A * B.T
 * where A is [M, K], B is [N, K], and C is [M, N]. All tensors are
 * expected to have row-major layout.
 *
 * @param M The variable dimension of matrix A and C.
 * @param A Pointer to the input matrix A [M, 4096] in device memory (dtype=half).
 * @param B Pointer to the input matrix B [4096, 4096] in device memory (dtype=half).
 * @param C Pointer to the output matrix C [M, 4096] in device memory (dtype=half).
 * @param stream The CUDA stream for kernel execution.
 */
void gemm_n4096_k4096_launcher(
    int M,
    const half* A,
    const half* B,
    half* C,
    cudaStream_t stream
);

#endif // GEMM_N4096_K4096_KERNEL_H
\end{minted}
\end{adjustwidth}

\texttt{kernel.cu}:\\

\captionsetup{type=listing}
\begin{adjustwidth}{0.5em}{0pt}
\begin{minted}{cpp}
#include "kernel.h"
#include <cublas_v2.h>
#include <cuda_fp16.h>

#include <stdexcept>
#include <string>

// Helper for checking cuBLAS status and converting it to a descriptive error
#define CUBLAS_CHECK(status)                                         \
  do {                                                               \
    cublasStatus_t err = (status);                                   \
    if (err != CUBLAS_STATUS_SUCCESS) {                              \
      throw std::runtime_error("cuBLAS error: " + std::to_string(err) \
                               + " at " __FILE__ ":" + std::to_string(__LINE__)); \
    }                                                                \
  } while (0)

// Manages a singleton cuBLAS handle for efficiency.
// This avoids the overhead of creating/destroying the handle on every call.
cublasHandle_t get_cublas_handle() {
    static struct CublasHandle {
        cublasHandle_t handle;
        CublasHandle() {
            CUBLAS_CHECK(cublasCreate(&handle));
            // Enable Tensor Core operations for peak performance on B200
            CUBLAS_CHECK(cublasSetMathMode(handle, CUBLAS_TENSOR_OP_MATH));
        }
        ~CublasHandle() {
            if (handle) cublasDestroy(handle);
        }
    } singleton_handle;
    return singleton_handle.handle;
}

/**
 * @brief Host-side launcher implementation using cuBLAS.
 */
void gemm_n4096_k4096_launcher(
    int M,
    const half* A,
    const half* B,
    half* C,
    cudaStream_t stream
) {
    cublasHandle_t handle = get_cublas_handle();
    CUBLAS_CHECK(cublasSetStream(handle, stream));

    const int N = 4096;
    const int K = 4096;

    const float alpha = 1.0f;
    const float beta = 0.0f;

    // The key to using cuBLAS (column-major) with row-major PyTorch tensors is
    // to rephrase the operation in a way that cuBLAS understands and that results
    // in the correct memory layout for the output.
    //
    // 1. Goal (Row-Major): C_rm[M, N] = A_rm[M, K] * B_rm.T[K, N]
    //
    // 2. cuBLAS View (Column-Major): cuBLAS interprets the memory of a row-major
    //    matrix X_rm[rows, cols] as a column-major matrix X_cm[cols, rows].
    //    - A_rm[M, K] is seen as A_cm[K, M].
    //    - B_rm[N, K] is seen as B_cm[K, N].
    //    - C_rm[M, N] is seen as C_cm[N, M].
    //
    // 3. Transformation: The equation C_rm = A_rm * B_rm.T is equivalent to
    //    C_cm.T = A_cm.T * (B_cm.T).T => C_cm.T = A_cm.T * B_cm.
    //    Taking the transpose of the whole equation gives us what cuBLAS should compute:
    //    C_cm = (A_cm.T * B_cm).T = B_cm.T * A_cm.
    //
    // 4. cuBLAS Call: We ask cuBLAS to compute D = op1 * op2, where the result D
    //    is written into the memory of C.
    //    - op1 = B_cm.T. This means the first matrix is B, and transa=CUBLAS_OP_T.
    //    - op2 = A_cm.   This means the second matrix is A, and transb=CUBLAS_OP_N.
    //
    // 5. Dimensions for cuBLAS:
    //    - m = rows of op1 (B.T) = N
    //    - n = cols of op2 (A)   = M
    //    - k = common dimension  = K
    //    The output matrix will be [m, n] = [N, M] in column-major layout, which
    //    perfectly matches the memory layout of our desired row-major C_rm[M, N].
    //    This resolves the illegal memory access and ensures correctness.
    const int lda = K; // Leading dimension of A_rm[M, K] is K
    const int ldb = K; // Leading dimension of B_rm[N, K] is K
    const int ldc = N; // Leading dimension of C_rm[M, N] is N

    CUBLAS_CHECK(cublasGemmEx(
        handle,
        CUBLAS_OP_T,        // transa: Corresponds to first matrix (B), transposed
        CUBLAS_OP_N,        // transb: Corresponds to second matrix (A), not transposed
        N,                  // m: rows of op(B.T)
        M,                  // n: columns of op(A)
        K,                  // k: common dimension
        &alpha,             // alpha
        B,                  // Pointer to the first matrix (B)
        CUDA_R_16F,         // Btype
        ldb,                // ldb (leading dimension of B)
        A,                  // Pointer to the second matrix (A)
        CUDA_R_16F,         // Atype
        lda,                // lda (leading dimension of A)
        &beta,              // beta
        C,                  // Pointer to C
        CUDA_R_16F,         // Ctype
        ldc,                // ldc (leading dimension of C)
        CUDA_R_32F,         // computeType: Use FP32 accumulators for precision
        CUBLAS_GEMM_DEFAULT_TENSOR_OP // algorithm: Use default heuristic for Tensor Cores
    ));
}
\end{minted}
\end{adjustwidth}

\texttt{main.cpp}:\\

\captionsetup{type=listing}
\begin{adjustwidth}{0.5em}{0pt}
\begin{minted}{cpp}
#include <torch/extension.h>
#include <c10/cuda/CUDAStream.h>

#include "kernel.h"

#include <stdexcept>
#include <string>

// Helper macros for concise tensor validation
#define CHECK_CUDA(x) TORCH_CHECK(x.is_cuda(), #x " must be a CUDA tensor")
#define CHECK_CONTIGUOUS(x) TORCH_CHECK(x.is_contiguous(), #x " must be contiguous")
#define CHECK_HALF(x) TORCH_CHECK(x.scalar_type() == torch::kFloat16, #x " must be a float16 tensor")

/**
 * @brief PyTorch extension entry point for the GEMM operation.
 *
 * This function validates input tensors and calls the CUDA kernel launcher
 * to perform the computation C = A * B.T on the GPU.
 *
 * @param A A torch::Tensor of shape [M, 4096] and dtype float16.
 * @param B A torch::Tensor of shape [4096, 4096] and dtype float16.
 * @return A torch::Tensor C of shape [M, 4096] and dtype float16 containing the result.
 */
torch::Tensor run(torch::Tensor A, torch::Tensor B) {
    // --- Input Validation ---
    CHECK_CUDA(A);
    CHECK_CUDA(B);
    CHECK_CONTIGUOUS(A);
    CHECK_CONTIGUOUS(B);
    CHECK_HALF(A);
    CHECK_HALF(B);

    TORCH_CHECK(A.dim() == 2, "A must be a 2D tensor");
    TORCH_CHECK(B.dim() == 2, "B must be a 2D tensor");

    // --- Dimension Checks ---
    const int M = A.size(0);
    const int K_A = A.size(1);
    const int N_B = B.size(0);
    const int K_B = B.size(1);

    const int N_spec = 4096;
    const int K_spec = 4096;

    TORCH_CHECK(K_A == K_spec, "A must have shape [M, 4096], but K is ", K_A);
    TORCH_CHECK(N_B == N_spec, "B must have shape [4096, 4096], but N is ", N_B);
    TORCH_CHECK(K_B == K_spec, "B must have shape [4096, 4096], but K is ", K_B);
    TORCH_CHECK(A.device() == B.device(), "Tensors must be on the same CUDA device");

    // --- Output Tensor Allocation ---
    auto C_options = torch::TensorOptions()
        .device(A.device())
        .dtype(A.scalar_type());
    auto C = torch::empty({M, N_spec}, C_options);

    // --- Kernel Execution ---
    try {
        // Get the current CUDA stream from PyTorch's context to ensure proper synchronization
        cudaStream_t stream = at::cuda::getCurrentCUDAStream();

        // Get raw data pointers. at::Half is compatible with cuda_fp16.h::half
        const half* A_ptr = reinterpret_cast<const half*>(A.data_ptr<at::Half>());
        const half* B_ptr = reinterpret_cast<const half*>(B.data_ptr<at::Half>());
        half* C_ptr = reinterpret_cast<half*>(C.data_ptr<at::Half>());

        // Launch the cuBLAS-based kernel
        gemm_n4096_k4096_launcher(M, A_ptr, B_ptr, C_ptr, stream);

    } catch (const std::exception& e) {
        // Propagate exceptions from the CUDA/cuBLAS calls to Python
        throw std::runtime_error(std::string("CUDA kernel execution failed: ") + e.what());
    }

    // Check for any asynchronous errors from the kernel launch. This is good practice.
    cudaError_t err = cudaGetLastError();
    if (err != cudaSuccess) {
        throw std::runtime_error(std::string("CUDA asynchronous error: ") + cudaGetErrorString(err));
    }
    return C;
}

// Pybind11 module definition to expose the 'run' function to Python
PYBIND11_MODULE(TORCH_EXTENSION_NAME, m) {
    m.def("run", &run, "GEMM (A * B.T) for N=4096, K=4096 using a B200-optimized cuBLAS kernel");
}
\end{minted}
\end{adjustwidth}

\clearpage

\subsection{Triton GQA Paged Decode Attention}
\label{appendix:triton_attention}

\begin{itemize}
  \item \textbf{Language} — Triton.
  \item \textbf{Source} — Generated by GPT-o3 from the \texttt{gqa\_paged\_decode\_h32\_kv8\_d128\_ps1} spec.
  \item \textbf{Correctness} — Correct on all evaluated workloads.
  \item \textbf{Performance (vs.\ baseline (FlashInfer))} — Speedup is \textbf{$0.19\times$} across all workload, with a best case of \textbf{$0.98\times$} on 3 workload.
  \item \textbf{Throughput (FLOPs/TFLOPs)} — Under grouped-query attention (group size $H_q/H_{kv}$), the dominant arithmetic for a decode step comprises
the $QK^\top$ dot products and the attention--$V$ product (softmax omitted as lower-order). Given wall-clock time $t$ seconds for the step,
\[
\mathrm{TFLOPs} \;=\; \frac{\mathrm{FLOPs}_{\text{step}}}{t \cdot 10^{12}}
\;=\; \frac{4\,H_q\,d\,\texttt{num\_kv\_indices}}{t \cdot 10^{12}}.
\]
\end{itemize}

\captionsetup{type=listing}
\begin{adjustwidth}{0.5em}{0pt}
\begin{minted}{python}
import math
import torch
import triton
import triton.language as tl


@triton.jit
def gqa_paged_decode_kernel(
    q_ptr,               # *bf16  [B, 32, 128]
    k_ptr,               # *bf16  [N_pages, 8, 128]  (page_size squeezed)
    v_ptr,               # *bf16  [N_pages, 8, 128]  (page_size squeezed)
    kv_indptr_ptr,       # *int32 [B + 1]
    kv_indices_ptr,      # *int32 [num_kv_indices]
    sm_scale,            # fp32 scalar
    out_ptr,             # *bf16  [B, 32, 128]
    lse_ptr,             # *fp32  [B, 32]
    BLOCK_T: tl.constexpr,
    HEAD_DIM: tl.constexpr,
    NUM_QO_HEADS: tl.constexpr,
    NUM_KV_HEADS: tl.constexpr,
):
    pid = tl.program_id(0)

    batch_idx = pid // NUM_QO_HEADS
    qo_head   = pid % NUM_QO_HEADS
    gqa_ratio = NUM_QO_HEADS // NUM_KV_HEADS
    kv_head   = qo_head // gqa_ratio

    # ---- strides (in elements, not bytes) ----
    stride_q_batch    = NUM_QO_HEADS * HEAD_DIM
    stride_q_head     = HEAD_DIM

    stride_k_page     = NUM_KV_HEADS * HEAD_DIM         # page_size = 1
    stride_k_kv_head  = HEAD_DIM

    stride_v_page     = stride_k_page
    stride_v_kv_head  = HEAD_DIM

    # ---- load query vector ----
    d_offs = tl.arange(0, HEAD_DIM)
    q_ptr_head = q_ptr + batch_idx * stride_q_batch + qo_head * stride_q_head + d_offs
    q_vec = tl.cast(tl.load(q_ptr_head), tl.float32)

    # ---- sequence token range ----
    start = tl.load(kv_indptr_ptr + batch_idx)
    end   = tl.load(kv_indptr_ptr + batch_idx + 1)
    num_tokens = end - start

    # ---- streaming softmax vars ----
    m_val   = tl.full([], -1e30, tl.float32)          # running max
    d_val   = tl.zeros([], tl.float32)                # running sum exp
    o_vec   = tl.zeros([HEAD_DIM], tl.float32)        # running output vector

    offset = tl.zeros([], tl.int32)

    while offset < num_tokens:
        t_offs      = tl.arange(0, BLOCK_T)
        remain      = num_tokens - offset
        tok_mask    = t_offs < remain

        # ---- load page indices ----
        pages = tl.load(kv_indices_ptr + start + offset + t_offs,
                        mask=tok_mask, other=0)

        # ---- gather K / V ----
        k_ptrs = k_ptr + pages[:, None] * stride_k_page + kv_head * stride_k_kv_head + d_offs[None, :]
        v_ptrs = v_ptr + pages[:, None] * stride_v_page + kv_head * stride_v_kv_head + d_offs[None, :]

        k_block = tl.cast(tl.load(k_ptrs, mask=tok_mask[:, None], other=0), tl.float32)
        v_block = tl.cast(tl.load(v_ptrs, mask=tok_mask[:, None], other=0), tl.float32)

        # ---- logits ----
        logits = tl.sum(k_block * q_vec[None, :], axis=1) * sm_scale
        logits = tl.where(tok_mask, logits, -1e30)

        # ---- block softmax ----
        m_block        = tl.max(logits, axis=0)
        exp_logits     = tl.exp(logits - m_block)
        sum_exp_block  = tl.sum(exp_logits, axis=0)
        weighted_v     = tl.sum(exp_logits[:, None] * v_block, axis=0)

        # ---- merge with running values ----
        new_m      = tl.maximum(m_val, m_block)
        alpha_prev = tl.exp(m_val - new_m)
        alpha_blk  = tl.exp(m_block - new_m)

        o_vec = o_vec * alpha_prev + weighted_v * alpha_blk
        d_val = d_val * alpha_prev + sum_exp_block * alpha_blk
        m_val = new_m

        offset += BLOCK_T

    inv_d   = tl.where(d_val == 0, 0.0, 1.0 / d_val)
    out_vec = o_vec * inv_d
    log2e   = 1.4426950408889634
    lse_val = tl.where(d_val == 0,
                       -1e30,
                       (tl.log(d_val) + m_val) * log2e)

    # ---- store ----
    out_ptr_head = out_ptr + batch_idx * stride_q_batch + qo_head * stride_q_head + d_offs
    tl.store(out_ptr_head, tl.cast(out_vec, tl.bfloat16))

    lse_ptr_head = lse_ptr + batch_idx * NUM_QO_HEADS + qo_head
    tl.store(lse_ptr_head, lse_val)


def run(q,
        k_cache,
        v_cache,
        kv_indptr,
        kv_indices,
        sm_scale: float | None = None):
    """
    Entry point for gqa_paged_decode_h32_kv8_d128_ps1.
    Returns (output, lse).
    """
    if sm_scale is None:
        sm_scale = 1.0 / math.sqrt(128.0)

    if not torch.cuda.is_available():
        raise RuntimeError("CUDA device is required to run Triton kernels.")

    # move tensors to GPU if necessary
    tensors = [q, k_cache, v_cache, kv_indptr, kv_indices]
    device_tensors = [t.cuda() if not t.is_cuda else t for t in tensors]
    q_dev, k_dev, v_dev, iptr_dev, idx_dev = [t.contiguous() for t in device_tensors]

    batch_size = q_dev.shape[0]
    num_qo_heads = 32
    head_dim = 128

    # squeeze page dimension (=1)
    k_dev_flat = k_dev.squeeze(1).contiguous()
    v_dev_flat = v_dev.squeeze(1).contiguous()

    out_dev = torch.empty((batch_size, num_qo_heads, head_dim),
                          dtype=torch.bfloat16,
                          device=q_dev.device)
    lse_dev = torch.empty((batch_size, num_qo_heads),
                          dtype=torch.float32,
                          device=q_dev.device)

    # launch kernel
    BLOCK_T = 128
    grid = (batch_size * num_qo_heads,)

    gqa_paged_decode_kernel[grid](
        q_dev, k_dev_flat, v_dev_flat,
        iptr_dev, idx_dev,
        sm_scale,
        out_dev, lse_dev,
        BLOCK_T=BLOCK_T,
        HEAD_DIM=128,
        NUM_QO_HEADS=32,
        NUM_KV_HEADS=8,
        num_warps=4,
        num_stages=4,
    )

    # move back to original device if needed
    if not q.is_cuda:
        return out_dev.cpu(), lse_dev.cpu()
    return out_dev, lse_dev
\end{minted}
\end{adjustwidth}

\clearpage

\subsection{CUDA GQA Paged Decode Attention}
\label{appendix:cuda_attention}

\begin{itemize}
  \item \textbf{Language} — CUDA.
  \item \textbf{Source} — Generated by GPT-5 from the \texttt{gqa\_paged\_decode\_h32\_kv8\_d128\_ps1} spec.
  \item \textbf{Implementation method} — It uses a straightforward implementation with 4 warps per block, processing one batch element and one KV head per block. The kernel implements streaming softmax with scalar FP32 operations.
  \item \textbf{Correctness} — Correct across all evaluated workloads.
  \item \textbf{Performance (vs.\ baseline (FlashInfer)} — Speedup is \textbf{$0.02\times$} across all workload, with a best case of \textbf{$1.02\times$} on one workload with batch size 1.
\end{itemize}

\begin{adjustwidth}{0.5em}{0pt}
\captionsetup{type=listing}
\begin{minted}{cuda}
#include "kernel.h"

#include <ATen/cuda/CUDAContext.h>
#include <cuda_bf16.h>
#include <cuda_runtime.h>
#include <math_constants.h>
#include <cmath>
#include <limits>
#include <cstdio>

// Error checking macro
#ifndef CUDA_CHECK
#define CUDA_CHECK(expr) do {                                     \
  cudaError_t __err = (expr);                                     \
  if (__err != cudaSuccess) {                                     \
    printf("CUDA Error %s at %s:%d\n", cudaGetErrorString(__err), \
           __FILE__, __LINE__);                                   \
  }                                                               \
} while (0)
#endif

// Warp utilities
static inline __device__ float warp_reduce_sum(float val) {
  unsigned mask = 0xffffffffu;
  val += __shfl_down_sync(mask, val, 16);
  val += __shfl_down_sync(mask, val, 8);
  val += __shfl_down_sync(mask, val, 4);
  val += __shfl_down_sync(mask, val, 2);
  val += __shfl_down_sync(mask, val, 1);
  return val;
}

static inline __device__ float warp_broadcast(float val, int src_lane) {
  unsigned mask = 0xffffffffu;
  return __shfl_sync(mask, val, src_lane);
}

// Load 4 BF16s as 4 floats (contiguous) from ptr[offset .. offset+3]
static inline __device__ void load_bf16x4_to_f32(
    const __nv_bfloat16* __restrict__ ptr,
    int offset,
    float out[4]) {
  out[0] = __bfloat162float(ptr[offset + 0]);
  out[1] = __bfloat162float(ptr[offset + 1]);
  out[2] = __bfloat162float(ptr[offset + 2]);
  out[3] = __bfloat162float(ptr[offset + 3]);
}

// Store 4 floats as BF16s to ptr[offset .. offset+3]
static inline __device__ void store_f32x4_to_bf16(
    __nv_bfloat16* __restrict__ ptr,
    int offset,
    const float in[4]) {
  ptr[offset + 0] = __float2bfloat16(in[0]);
  ptr[offset + 1] = __float2bfloat16(in[1]);
  ptr[offset + 2] = __float2bfloat16(in[2]);
  ptr[offset + 3] = __float2bfloat16(in[3]);
}

template <int kBlockThreads>
__launch_bounds__(kBlockThreads, 2)
__global__ void gqa_paged_decode_h32_kv8_d128_ps1_kernel(
    const __nv_bfloat16* __restrict__ q,          // [B, 32, 128]
    const __nv_bfloat16* __restrict__ k_cache,    // [num_pages, 1, 8, 128] -> flat [num_pages*8, 128]
    const __nv_bfloat16* __restrict__ v_cache,    // [num_pages, 1, 8, 128] -> flat [num_pages*8, 128]
    const int32_t* __restrict__ kv_indptr,        // [B+1]
    const int32_t* __restrict__ kv_indices,       // [num_kv_indices]
    float sm_scale,
    __nv_bfloat16* __restrict__ out,              // [B, 32, 128]
    float* __restrict__ lse_out,                  // [B, 32]
    int num_batches,
    int num_pages_total
) {
  // Block mapping:
  // grid.x = batch index
  // grid.y = kv_head index in [0, 8)
  const int b = blockIdx.x;
  const int kv_head = blockIdx.y; // 0..7

  if (b >= num_batches || kv_head >= kNumKVHeads) {
    return;
  }

  // Thread mapping:
  const int tid = threadIdx.x;       // 0..127
  const int warp_id = tid >> 5;      // 0..3 (4 warps per block)
  const int lane_id = tid & 31;      // 0..31

  // The 4 query heads attached to this KV head
  const int q_head = kv_head * kGQARatio + warp_id; // 0..31

  // Pointers advance helpers
  const int q_stride_h = kHeadDim;
  const int q_stride_head = kNumQOHeads * kHeadDim;

  // Input sequence token range for this batch item
  const int32_t page_start = kv_indptr[b];
  const int32_t page_end   = kv_indptr[b + 1];
  const int32_t num_tokens = page_end - page_start;

  // Shared buffers for one token's K and V vector for this kv_head
  extern __shared__ float smem[];
  float* sh_k = smem;                // [128]
  float* sh_v = smem + kHeadDim;     // [128]
  __shared__ int s_page;

  // Preload Q (each warp for its own q_head)
  // Each lane holds 4 elements to cover 128 dims: 32 lanes * 4 = 128
  const int q_base_offset = (b * q_stride_head) + (q_head * q_stride_h);
  const int d_base = lane_id * 4;

  float q_reg[4];
  // Safe load even if num_tokens == 0
  load_bf16x4_to_f32(q + q_base_offset, d_base, q_reg);

  // Accumulators per warp/head
  float out_acc[4] = {0.f, 0.f, 0.f, 0.f};
  float m = -CUDART_INF_F;  // running max of logits (scaled)
  float s = 0.f;            // running sum of exp(logit - m)

  // If no tokens: write zeros and lse = -inf and return
  if (num_tokens <= 0) {
    float zeros[4] = {0.f, 0.f, 0.f, 0.f};
    store_f32x4_to_bf16(out + (b * kNumQOHeads + q_head) * kHeadDim, d_base, zeros);
    if (lane_id == 0) {
      lse_out[b * kNumQOHeads + q_head] = -CUDART_INF_F;
    }
    return;
  }

  // Iterate over tokens
  for (int t = 0; t < num_tokens; ++t) {
    if (tid == 0) {
      s_page = kv_indices[page_start + t];
    }
    __syncthreads();

    // Bounds check for safety (though constraints guarantee validity)
    int page_id = s_page;
    if (page_id < 0) page_id = 0;
    if (page_id >= num_pages_total) page_id = (num_pages_total - 1);

    // Flattened (page_size=1): [num_pages, 1, 8, 128] -> [num_pages*8, 128]
    // Base index for this token and kv_head
    size_t base_idx = (static_cast<size_t>(page_id) * kNumKVHeads + kv_head) * kHeadDim;

    // Cooperatively load K and V vectors into shared memory as float
    if (tid < kHeadDim) {
      sh_k[tid] = __bfloat162float(k_cache[base_idx + tid]);
      sh_v[tid] = __bfloat162float(v_cache[base_idx + tid]);
    }
    __syncthreads();

    // Each warp computes its logit: dot(q, k) using 4 elements per lane
    float partial = 0.f;
    partial += q_reg[0] * sh_k[d_base + 0];
    partial += q_reg[1] * sh_k[d_base + 1];
    partial += q_reg[2] * sh_k[d_base + 2];
    partial += q_reg[3] * sh_k[d_base + 3];

    float sum = warp_reduce_sum(partial);
    float logit = warp_broadcast(sum, 0) * sm_scale;

    // Streaming softmax update
    float m_new = fmaxf(m, logit);
    float e1 = __expf(m - m_new);         // scale for previous accumulators
    float e2 = __expf(logit - m_new);     // new contribution

    s = s * e1 + e2;
    // Update vector accumulator
    out_acc[0] = out_acc[0] * e1 + e2 * sh_v[d_base + 0];
    out_acc[1] = out_acc[1] * e1 + e2 * sh_v[d_base + 1];
    out_acc[2] = out_acc[2] * e1 + e2 * sh_v[d_base + 2];
    out_acc[3] = out_acc[3] * e1 + e2 * sh_v[d_base + 3];

    m = m_new;

    __syncthreads();
  }

  // Finalize: normalize output by s, write lse base-2
  float inv_s = 1.f / s;
  float out_final[4] = {
    out_acc[0] * inv_s,
    out_acc[1] * inv_s,
    out_acc[2] * inv_s,
    out_acc[3] * inv_s
  };

  // Store output
  store_f32x4_to_bf16(out + (b * kNumQOHeads + q_head) * kHeadDim, d_base, out_final);

  // lse = logsumexp(logits_scaled) base 2 = (log(s) + m) / ln(2)
  if (lane_id == 0) {
    constexpr float ln2 = 0.693147180559945309417232121458176568f;
    float lse_val = (logf(s) + m) / ln2;
    lse_out[b * kNumQOHeads + q_head] = lse_val;
  }
}

// Host wrapper: validate inputs, set up launch config, and launch kernel
void gqa_paged_decode_h32_kv8_d128_ps1_cuda(
    const torch::Tensor& q,            // [B, 32, 128] bfloat16
    const torch::Tensor& k_cache,      // [num_pages, 1, 8, 128] bfloat16
    const torch::Tensor& v_cache,      // [num_pages, 1, 8, 128] bfloat16
    const torch::Tensor& kv_indptr,    // [B+1] int32
    const torch::Tensor& kv_indices,   // [num_kv_indices] int32
    float sm_scale,
    torch::Tensor& output,             // [B, 32, 128] bfloat16
    torch::Tensor& lse                 // [B, 32] float32
) {
  TORCH_CHECK(q.is_cuda(), "q must be CUDA tensor");
  TORCH_CHECK(k_cache.is_cuda(), "k_cache must be CUDA tensor");
  TORCH_CHECK(v_cache.is_cuda(), "v_cache must be CUDA tensor");
  TORCH_CHECK(kv_indptr.is_cuda(), "kv_indptr must be CUDA tensor");
  TORCH_CHECK(kv_indices.is_cuda(), "kv_indices must be CUDA tensor");
  TORCH_CHECK(output.is_cuda(), "output must be CUDA tensor");
  TORCH_CHECK(lse.is_cuda(), "lse must be CUDA tensor");

  TORCH_CHECK(q.dim() == 3, "q must be [B, 32, 128]");
  TORCH_CHECK(q.size(1) == kNumQOHeads && q.size(2) == kHeadDim,
              "q must be [B, 32, 128]");
  TORCH_CHECK(q.scalar_type() == at::kBFloat16, "q must be bfloat16");

  TORCH_CHECK(k_cache.dim() == 4, "k_cache must be [num_pages, 1, 8, 128]");
  TORCH_CHECK(k_cache.size(1) == 1 && k_cache.size(2) == kNumKVHeads && k_cache.size(3) == kHeadDim,
              "k_cache must be [num_pages, 1, 8, 128]");
  TORCH_CHECK(k_cache.scalar_type() == at::kBFloat16, "k_cache must be bfloat16");

  TORCH_CHECK(v_cache.dim() == 4, "v_cache must be [num_pages, 1, 8, 128]");
  TORCH_CHECK(v_cache.size(1) == 1 && v_cache.size(2) == kNumKVHeads && v_cache.size(3) == kHeadDim,
              "v_cache must be [num_pages, 1, 8, 128]");
  TORCH_CHECK(v_cache.scalar_type() == at::kBFloat16, "v_cache must be bfloat16");

  TORCH_CHECK(kv_indptr.dim() == 1, "kv_indptr must be 1D");
  TORCH_CHECK(kv_indptr.scalar_type() == at::kInt, "kv_indptr must be int32");
  TORCH_CHECK(kv_indices.dim() == 1, "kv_indices must be 1D");
  TORCH_CHECK(kv_indices.scalar_type() == at::kInt, "kv_indices must be int32");

  const int64_t B = q.size(0);
  TORCH_CHECK(kv_indptr.size(0) == B + 1, "len_indptr must be batch_size + 1");

  const int64_t num_pages = k_cache.size(0);
  TORCH_CHECK(output.sizes() == q.sizes(), "output must be same shape as q");
  TORCH_CHECK(output.scalar_type() == at::kBFloat16, "output must be bfloat16");
  TORCH_CHECK(lse.dim() == 2 && lse.size(0) == B && lse.size(1) == kNumQOHeads,
              "lse must be [B, 32]");
  TORCH_CHECK(lse.scalar_type() == at::kFloat, "lse must be float32");

  // Launch config
  dim3 grid;
  grid.x = static_cast<unsigned>(B);
  grid.y = static_cast<unsigned>(kNumKVHeads);
  grid.z = 1;

  constexpr int kBlockThreads = 128; // 4 warps per block => 4 Q heads per KV head group
  dim3 block(kBlockThreads);

  // Shared memory for K and V: 2 * 128 floats
  size_t shmem_bytes = 2 * kHeadDim * sizeof(float);

  auto stream = at::cuda::getCurrentCUDAStream();

  const __nv_bfloat16* q_ptr = reinterpret_cast<const __nv_bfloat16*>(q.data_ptr<at::BFloat16>());
  const __nv_bfloat16* k_ptr = reinterpret_cast<const __nv_bfloat16*>(k_cache.data_ptr<at::BFloat16>());
  const __nv_bfloat16* v_ptr = reinterpret_cast<const __nv_bfloat16*>(v_cache.data_ptr<at::BFloat16>());
  const int32_t* indptr_ptr = kv_indptr.data_ptr<int32_t>();
  const int32_t* indices_ptr = kv_indices.data_ptr<int32_t>();
  __nv_bfloat16* out_ptr = reinterpret_cast<__nv_bfloat16*>(output.data_ptr<at::BFloat16>());
  float* lse_ptr = lse.data_ptr<float>();

  gqa_paged_decode_h32_kv8_d128_ps1_kernel<kBlockThreads><<<grid, block, shmem_bytes, stream>>>(
      q_ptr, k_ptr, v_ptr, indptr_ptr, indices_ptr, sm_scale, out_ptr, lse_ptr,
      static_cast<int>(B), static_cast<int>(num_pages)
  );

  CUDA_CHECK(cudaGetLastError());
}
\end{minted}
\end{adjustwidth}

\end{adjustwidth}

\section{Solution Generation Prompt}
This appendix contains the prompts used to generate the kernel solutions in the dataset and case studies. Rather than providing prescriptive kernel optimization advice, these prompts focus on generating syntactically correct, parsable code while allowing the agent to explore implementation strategies autonomously. The base prompt is used for initial kernel proposal, and the optimization prompt for iterative improvement.

\subsection{Triton Base Prompt}
\label{subsec:triton-base-prompt}

\begin{minted}{text}
Generate a Triton kernel optimized for {target_gpu} GPU for

{definition}

Triton Version: 3.3.1

Requirements:
- Write clean, efficient Triton code optimized for {target_gpu} architecture
- Use modern Triton syntax with proper grid computation and language features
- Include necessary imports (torch, triton, triton.language as tl)
- Implement the exact functionality described in the specification
- The reference code provides the mathematical specification but is unoptimized - your Triton implementation should match its computational accuracy while delivering high performance
- Use the definition's tensor shapes, dtypes, and axes information to guide memory access patterns and optimization strategies
- Optimize for {target_gpu} GPU characteristics (memory hierarchy, compute units, etc.)

The wrapper function MUST handle complete device management:
- Move CPU tensors to GPU if needed (use .to('cuda') or .cuda() when torch.cuda.is_available())
- Raise clear errors if CUDA is not available for GPU tensors
- Call the triton kernel with GPU tensors
- Move results back to original device of input tensors
- Handle both args and kwargs properly
- Preserve original tensor devices and restore them for outputs

IMPORTANT: Use only valid Python/Triton syntax:
- NO hexadecimal float literals (0x1.234p5) - use decimal equivalents
- NO C/CUDA specific syntax - this is Python/Triton code
- All code must be valid Python that passes ast.parse()

- Expose a "run" entry point function that can be called to execute the kernel
- Return only the code, no explanations or markdown formatting

Generate complete, runnable code only - no framework will add device handling wrapper code.

Generate the implementation:
\end{minted}

\subsection{Triton Optimization Prompt}
\label{subsec:triton-optimization-prompt}

\begin{minted}{text}
You are optimizing a Triton kernel for {target_gpu} GPU. The current implementation has issues that need to be fixed.

Original Specification:
{definition}

Current Implementation Status:
{trace_logs}

Current Implementation:
{current_code}

Optimization Strategy:
1. ENSURE CORRECTNESS: If there are compile errors, runtime errors, or incorrect outputs, focus entirely on fixing these issues
   - Analyze compilation errors and fix syntax/API usage
   - Fix runtime errors like shape mismatches, memory access violations
   - Ensure numerical correctness matches the reference implementation

2. OPTIMIZE PERFORMANCE: if the current kernel is functionally correct, focus on performance optimizations
   - Optimize memory access patterns for {target_gpu}
   - Tune block sizes and grid dimensions
   - Use appropriate Triton language features for vectorization
   - Minimize global memory transactions

Requirements for the optimized implementation:
- Write clean, efficient Triton code optimized for {target_gpu} architecture
- Use modern Triton syntax with proper grid computation and language features
- Include necessary imports (torch, triton, triton.language as tl)
- Fix all identified issues from the feedback
- Maintain or improve computational accuracy
- Preserve the same function signature and device handling as specified

The wrapper function MUST handle complete device management:
- Move CPU tensors to GPU if needed (use .to('cuda') or .cuda() when torch.cuda.is_available())
- Raise clear errors if CUDA is not available for GPU tensors
- Call the triton kernel with GPU tensors
- Move results back to original device of input tensors
- Handle both args and kwargs properly
- Preserve original tensor devices and restore them for outputs

IMPORTANT: Use only valid Python/Triton syntax:
- NO hexadecimal float literals (0x1.234p5) - use decimal equivalents
- NO C/CUDA specific syntax - this is Python/Triton code
- All code must be valid Python that passes ast.parse()

- Expose a "run" entry point function that can be called to execute the kernel
- Return only the improved code, no explanations or markdown formatting

Generate the corrected and optimized implementation:
\end{minted}

\subsection{CUDA Base Prompt}
\label{subsec:cuda-base-prompt}

\begin{minted}{text}
You are a code generator. Generate a CUDA kernel implementation optimized for {target_gpu} GPU for the following specification.

Specification:
{definition}

Requirements:
- Write clean, efficient CUDA C++ code optimized for {target_gpu} architecture
- Use proper CUDA syntax and memory management optimized for {target_gpu}
- Implement the exact functionality described in the specification
- The reference code provides the mathematical specification but is unoptimized - your CUDA implementation should match its computational accuracy while delivering high performance
- Use the definition's tensor shapes, dtypes, and axes information to guide memory access patterns and optimization strategies
- Optimize for {target_gpu} GPU characteristics (memory hierarchy, compute units, etc.)
- For fixed axis values, optimize specifically for those constants rather than general cases

IMPORTANT: Generate code in XML format with exactly 3 files with these strict names:

<header_file name="kernel.h">
- All CUDA kernel function declarations
- Host function declarations
- Any necessary struct/type definitions
- Include guards and necessary headers
</header_file>

<cuda_file name="kernel.cu">
- All __global__ kernel implementations
- All __device__ helper functions
- CUDA-specific optimizations and memory patterns
- Proper error checking and memory management
</cuda_file>

<cpp_file name="main.cpp">
- Host function that launches kernels
- Memory allocation and data transfer management
- Device management and error handling
- Entry point function named "run" that can be called to execute the implementation
- Handle both args and kwargs properly
- Move CPU data to GPU, execute kernels, and return results to CPU
- Include PyTorch C++ extension bindings using PYBIND11_MODULE
- The "run" function must be exposed to Python through the binding
- Include proper tensor type conversion between PyTorch tensors and CUDA pointers
- Include all necessary PyTorch headers: #include <torch/extension.h>
</cpp_file>

Code Generation Guidelines:
- Use modern CUDA features appropriate for {target_gpu}
- Optimize memory coalescing and reduce bank conflicts
- Utilize shared memory effectively for data reuse
- Consider occupancy and register usage
- Implement proper error checking with cudaGetLastError()
- Use appropriate grid and block dimensions for the problem size
- Leverage constant memory for frequently accessed read-only data
- Use PyTorch tensor API (torch::Tensor) for all tensor arguments in the "run" function
- Convert PyTorch tensors to CUDA pointers using .data_ptr<T>() with appropriate type (e.g., float, double, int)
- Ensure proper CUDA stream synchronization and error handling

Generate the implementation:
\end{minted}

\subsection{CUDA Optimization Prompt}
\label{subsec:cuda-optimization-prompt}

\begin{minted}{text}
You are optimizing a CUDA kernel for {target_gpu} GPU. The current implementation has issues that need to be fixed.

Original Specification:
{definition}

Current Implementation Status:
{trace_logs}

Current Implementation:
{current_code}

Optimization Strategy:
1. ENSURE CORRECTNESS: If there are compile errors, runtime errors, or incorrect outputs, focus entirely on fixing these issues
   - Analyze compilation errors and fix syntax/API usage
   - Fix runtime errors like shape mismatches, memory access violations, kernel launch failures
   - Ensure numerical correctness matches the reference implementation
   - Verify proper CUDA memory management and synchronization

2. OPTIMIZE PERFORMANCE: if the current kernel is functionally correct, focus on performance optimizations
   - Optimize memory access patterns and coalescing for {target_gpu}
   - Tune block sizes and grid dimensions for maximum occupancy
   - Utilize shared memory effectively to reduce global memory transactions
   - Optimize register usage and minimize divergent branches
   - Consider using specialized libraries if beneficial
   - Leverage constant axis values for compile-time optimizations

Requirements for the optimized implementation:
- Write clean, efficient CUDA C++ code optimized for {target_gpu} architecture
- Use proper CUDA syntax and modern features appropriate for {target_gpu}
- Fix all identified issues from the feedback
- Maintain or improve computational accuracy
- Preserve the same function signatures and device handling as specified
- For fixed axis values, optimize specifically for those constants rather than general cases

IMPORTANT: Generate code in XML format with exactly 3 files with these strict names:

<header_file name="kernel.h">
- All CUDA kernel function declarations
- Host function declarations
- Any necessary struct/type definitions
- Include guards and necessary headers
</header_file>

<cuda_file name="kernel.cu">
- All __global__ kernel implementations
- All __device__ helper functions
- CUDA-specific optimizations and memory patterns
- Proper error checking and memory management
</cuda_file>

<cpp_file name="main.cpp">
- Host function that launches kernels
- Memory allocation and data transfer management
- Device management and error handling
- Entry point function named "run" that can be called to execute the implementation
- Handle both args and kwargs properly
- Move CPU data to GPU, execute kernels, and return results to CPU
- Include PyTorch C++ extension bindings using PYBIND11_MODULE
- The "run" function must be exposed to Python through the binding
- Include proper tensor type conversion between PyTorch tensors and CUDA pointers
- Include all necessary PyTorch headers: #include <torch/extension.h>
</cpp_file>

Code Generation Guidelines:
- Use modern CUDA features appropriate for {target_gpu}
- Optimize memory coalescing and reduce bank conflicts
- Utilize shared memory effectively for data reuse
- Consider occupancy and register usage
- Implement proper error checking with cudaGetLastError()
- Use appropriate grid and block dimensions for the problem size
- Leverage constant memory for frequently accessed read-only data
- Use PyTorch tensor API (torch::Tensor) for all tensor arguments in the "run" function
- Convert PyTorch tensors to CUDA pointers using .data_ptr<T>() with appropriate type (e.g., float, double, int)
- Ensure proper CUDA stream synchronization and error handling

Generate the corrected and optimized implementation:
\end{minted}

\end{document}